\documentclass[10pt,twocolumn,letterpaper]{article}

\usepackage{multirow}
\usepackage{graphicx}
\usepackage{booktabs}
\usepackage{amsmath}
\usepackage{textcomp}
\usepackage{colortbl}
\usepackage{bbding}
\usepackage{footmisc}  
\renewcommand{\thefootnote}{\fnsymbol{footnote}}  

\usepackage[pagenumbers]{cvpr} 

\definecolor{cvprblue}{rgb}{0.21,0.49,0.74}
\usepackage[pagebackref,breaklinks,colorlinks,allcolors=cvprblue]{hyperref}

\def\paperID{12208} 
\def\confName{CVPR}
\def\confYear{2025}

\title{Robust Multimodal Survival Prediction with Conditional Latent Differentiation Variational AutoEncoder}

\author{
    Junjie Zhou, \; 
    Jiao Tang, \; 
    Yingli Zuo, \;
    Peng Wan, \;
    Daoqiang Zhang, \
    Wei Shao\footnotemark[1] \\
    {The College of Artificial Intelligence, Nanjing University of Aeronautics and Astronautics}\\ 
    {The Key Laboratory of Brain-Machine Intelligence Technology, Ministry of Education}\\ 
    {\tt\small junjiezhou@nuaa.edu.cn,~dqzhang@nuaa.edu.cn,~shaowei20022005@nuaa.edu.cn}
    \vspace{-2mm}
}

\begin{document}
\maketitle
\footnotetext[1]{Corresponding author}

\begin{abstract}
The integrative analysis of histopathological images and genomic data has received increasing attention for survival prediction of human cancers. However, the existing studies always hold the assumption that full modalities are available. As a matter of fact, the cost for collecting genomic data is high, which sometimes makes genomic data unavailable in testing samples. A common way of tackling such incompleteness is to generate the genomic representations from the pathology images. Nevertheless, such strategy still faces the following two challenges: (1) The gigapixel whole slide images (WSIs) are huge and thus hard for representation. (2) It is difficult to generate the genomic embeddings with diverse function categories in a unified generative framework. To address the above challenges, we propose a Conditional Latent Differentiation Variational AutoEncoder (LD-CVAE) for robust multimodal survival prediction, even with missing genomic data. Specifically, a Variational Information Bottleneck Transformer (VIB-Trans) module is proposed to learn compressed pathological representations from the gigapixel WSIs. To generate different functional genomic features, we develop a novel Latent Differentiation Variational AutoEncoder (LD-VAE) to learn the genomic and function-specific posteriors for the genomic embeddings with diverse functions. Finally, we use the product-of-experts technique to integrate the genomic posterior and image posterior for the joint latent distribution estimation in LD-CVAE. We test the effectiveness of our method on five different cancer datasets, and the experimental results demonstrate its superiority in both complete and missing modality scenarios. 
The code is released \footnote[2]{\href{https://github.com/JJ-ZHOU-Code/RobustMultiModel}{https://github.com/JJ-ZHOU-Code/RobustMultiModel}}.
\end{abstract}

\vspace{-3mm}
\section{Introduction}
\label{sec:intro}

\begin{figure}
    \centering
    \includegraphics[width=1.0\linewidth]{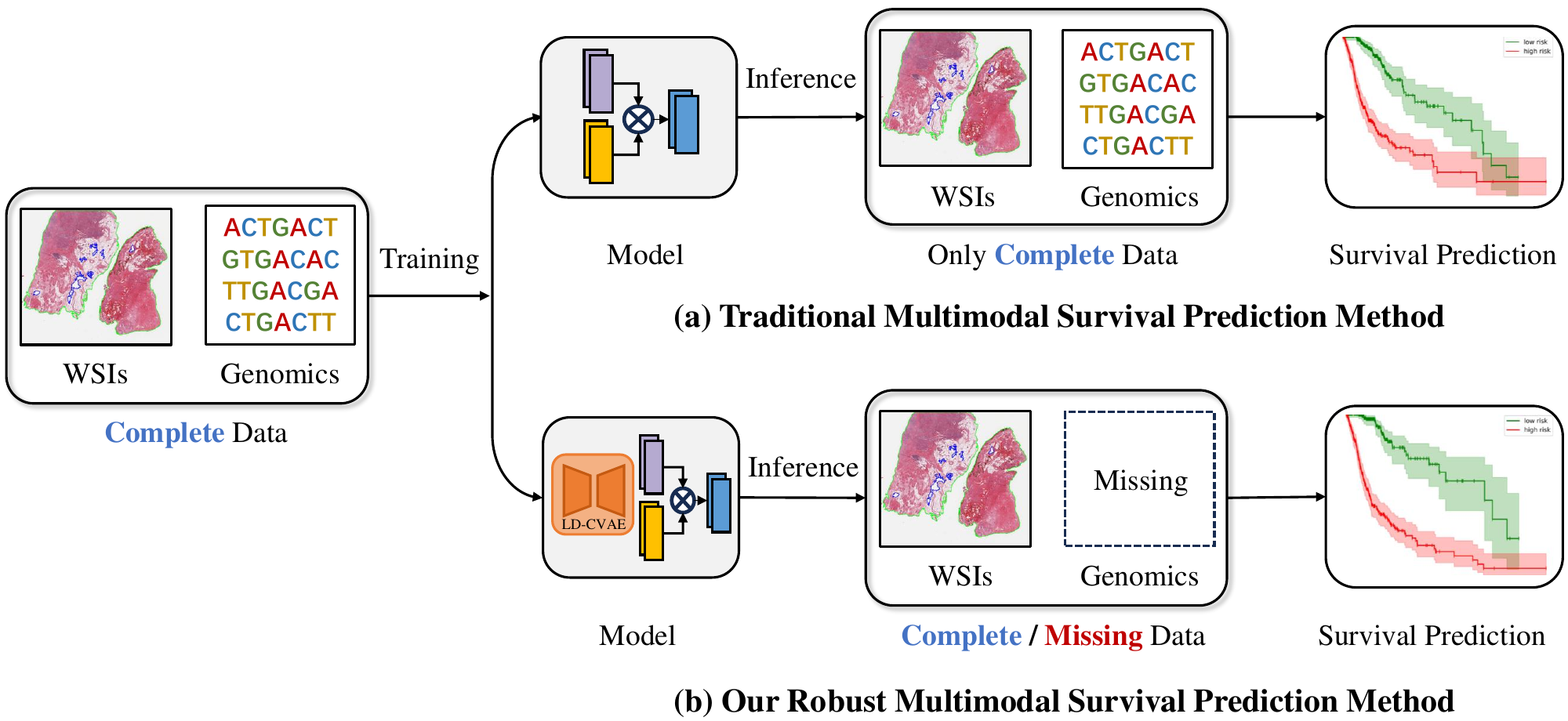}
    \caption{Comparison between (a) traditional and (b) our proposed multimodal survival prediction methods, where the traditional methods require the complete multi-modal data while our method is still effective if the genomic data is missing.}
    \label{fig:intro}
    \vspace{-4mm}
\end{figure}

Survival analysis aims at estimating the survival time for an individual or a group of patients, which is a valid solution for cancer treatment and precision medicine \cite{nagy2021pancancer, gyHorffy2021survival}. So far, diverse biomarkers are verified to be sensitive to the survival prediction of human cancers, and these biomarkers can be broadly classified into two categories, \emph{i.e.,} image biomarkers and genomic biomarkers \cite{acosta2022multimodal,steyaert2023multimodal}. Among all image biomarkers, the histopathological images provide rich visual information for the assessment of histological features, cellular organization, and other phenotypic characteristics that are generally considered as the golden standard for cancer prognosis \cite{kumar2020whole, lu2021data}. Besides histopathological images, genomic data also offers insights into the underlying molecular mechanisms of cancer \cite{zeng2023clinical,el2024regression}, and thus many researchers leverage patients' molecular profiles such as genetic alterations and gene expression signatures to drive prognostic practice \cite{ushijima2021mapping, muller2020landscape}. 

In many clinical and research studies, it is common to acquire multiple biomarkers for a more accurate assessment of disease status and progression stages \cite{esteva2022prostate}. Recently, researches also explore to combine both pathological images and genomic markers for the prognosis of cancer \cite{chen2020pathomic, chen2021multimodal, chen2022pan}. For instance, the authors in \cite{zhou2023cross,chen2021multimodal} have presented different integrative survival prediction models based on the co-attention transformer framework. Zuo \emph{et al.} \cite{zuo2022identify} explore the interactions among different tumor micro-environment components and propose an explainable multi-modal fusion framework for prognosis predictions of breast cancer. All these studies suggest that the combination of genomic data and pathological images can better stratify cancer patients with distinct prognosis than using single biomarker.

Although much progress has been achieved, the existing multimodal studies hold the assumption that both the imaging and genomic data are available. As a matter of fact, the cost for collecting genomic data is much higher than acquiring the pathology images \cite{payne2018cost}, which sometimes makes genomic data unavailable in the testing samples and thus deteriorates the prediction performance for multimodal learning. A common way to address the issue of missing modality is to utilize one modality to reconstruct the features of another modality, and thus the reconstructed feature can fill in for the missing ones \cite{ma2021smil, wu2018multimodal, ma2022multimodal, lee2023multimodal,wang2023multi, dorent2019hetero, havaei2016hemis}. However, such strategy still faces the following two challenges for the generation of the missing genomic features from the whole-slide pathological images (WSIs):

1. \emph{The representation of gigapixel whole slide images (WSIs)}: The WSIs are usually with huge size (\emph{e.g.,} $150,000$ $\times$ $150,000$ pixels) but only a small fraction of areas are closely associated with the genotype information \cite{campanella2019clinical, zhu2017wsisa}. It is challenge to learn effective representation from the WSIs for the reconstruction of the genomic feature.   

2. \emph{Generate the genomic embeddings with diverse function categories}: Since different genes are usually with diverse biological functional impact (\emph{e.g.,} oncogenesis and cell differentiation), it is difficult to address such difference for the generation of genomic embeddings in an unified generative framework.

To address the above challenges, we present a Conditional Latent Differentiation Variational AutoEncoder (LD-CVAE) that can reconstruct the genomic embeddings with diverse biological functions from WSIs for robust multimodal survival prediction, even with the missing genomic data (as shown in \cref{fig:intro}). The experimental results on five cancer cohorts derived from the The Cancer Genome Atlas (TCGA) clearly demonstrate its advantage for both complete and missing modality scenarios when comparing with the existing studies. We summarize the main contributions of this study as follows: 
\begin{itemize}
\item We introduce a robust multimodal survival prediction model that can handle the missing of the genomic data.
\item We develop a Variational Information Bottleneck Transformer (VIB-Trans), which employs both the information bottleneck theory and Transformer to obtain useful pathological representation for genomic profile reconstruction.
\item We propose a novel Latent Differentiation Variational AutoEncoder (LD-VAE) that can help generate different functional genomic embeddings by learning both the genomic and function-specific posteriors.
\item We further learn the joint posterior in LD-CVAE by the product-of-experts technique that can integrate genomic posterior and pathological posterior. Additionally, an alignment loss is introduced to align pathological and genomic latent distributions for better estimating the joint distribution.
\end{itemize}

\section{Related Work}
\label{sec:related}

\subsection{Multimodal Survival Analysis}
By integrating pathological WSIs and genomic data, multimodal approaches \cite{chen2020pathomic, chen2021multimodal, chen2022pan, xu2023multimodal, zhou2023cross, jaume2024modeling, zhang2024prototypical} provide a more comprehensive perspective on patient stratification and prognosis. For instance, Chen \emph{et al.} \cite{chen2020pathomic} 
use the Kronecker Product to model pairwise feature interactions across multi-modal data. 
Chen \emph{et al.} \cite{chen2021multimodal} present a multimodal co-Attention transformer (MCAT) framework
that learns an interpretable, dense co-attention mapping between WSIs and genomic features formulated in an embedding space. Furthermore, Xu \emph{et al.} \cite{xu2023multimodal} propose an optimal transport based co-attention learning module that can match instances between histology and genomics to represent the gigapixel WSI. Zhang \emph{et al.} \cite{zhang2024prototypical} explore multimodal cancer survival prediction and propose a new framework called PIBD aimed at addressing both “intra-model redundancy” and “inter-model redundancy” challenges. However, the above multimodal studies usually require the completeness of all modalities for training and evaluation, limiting their applicability in real-world with missing-modality challenges.
\subsection{Missing Modality in Multimodal Learning}
To address the missing modality issues in multimodal learning. Ma \emph{et al.} \cite{ma2021smil} propose a new method named SMIL that leverages Bayesian meta-learning for the generation of missing modality data.  Wu \emph{et al.} \cite{wu2018multimodal} introduce a multimodal variational autoencoder (MVAE) that uses a sub-sampled training paradigm to solve the missing modality problem. In addition, the study in \cite{ma2022multimodal} comprehensively examines the behavior of Transformers in the presence of modality-incomplete data and further improves the robustness of the Transformer via multi-task optimization. ShaSpec \cite{wang2023multi} is introduced to model and fuse shared and specific features to address the missing modality data. However, it is still challenge to recover the missing genomic profiles from WSIs since they are with high-resolution that are hard for representation. Additionally, the extracted genomic features have diverse biological functions, and it is not easy to reconstruct them in an unified framework.

\begin{figure*}
    \centering
    \includegraphics[width=1.0\linewidth]{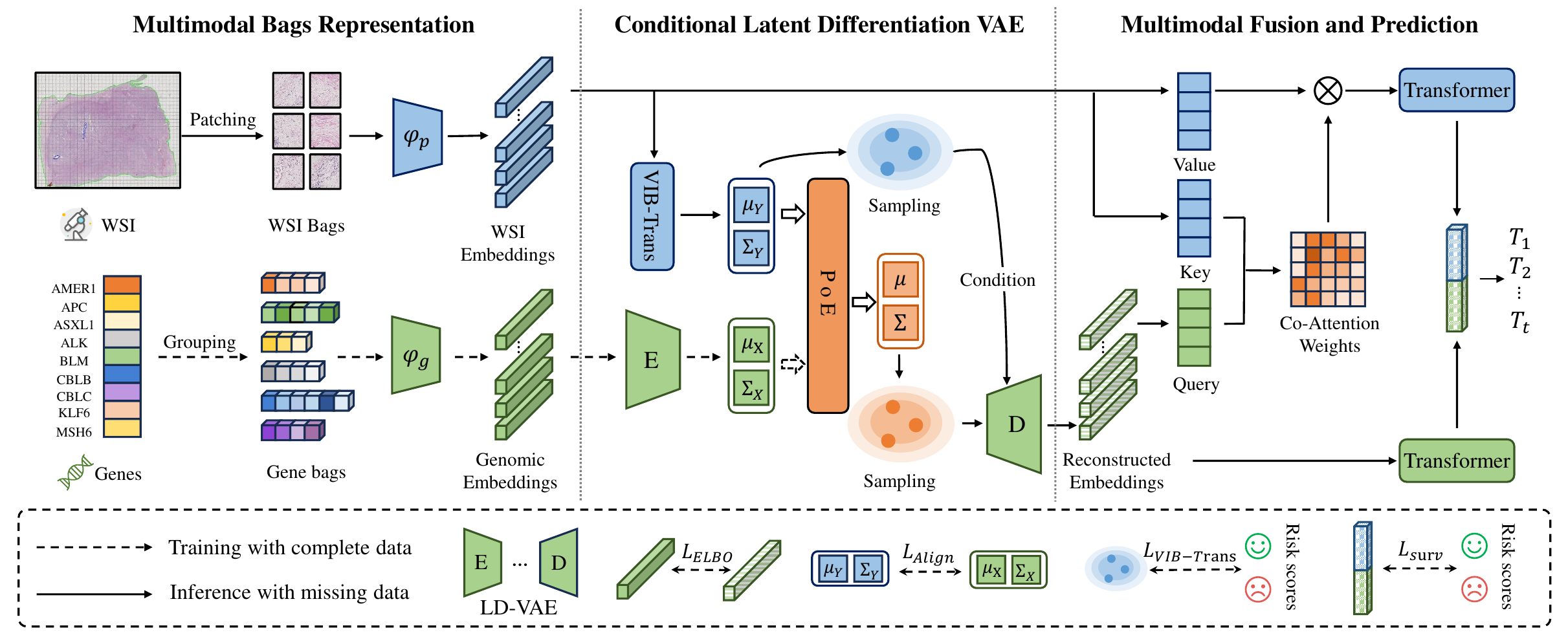}
    \caption{ The flowchart of the proposed robust multimodal survival prediction model is consisted of three steps. In the multimodal bag representation step, we extract multimodal features from both WSI and genomic data. Then, a conditional latent Differentiation Variational AutoEncoder (LD-CVAE) is proposed to reconstruct the genomic representation from the image data, and thus our method is still effective if the genomic data is missing in the testing stage.  Finally, the co-attention module is applied to guide the selection of survival-associated instance, which is then combined with the reconstructed genomic features for survival prediction.}
    \label{fig:overview}
    \vspace{-2mm}
\end{figure*}

\section{Method}
\label{sec:method}

\subsection{Overview}
\label{sec:method_overview}
The overall framework of the proposed robust multimodal survival prediction model is shown in \cref{fig:overview}. Specifically, after extracting the multimodal features from both WSI and genomic data, a VIB-Trans block is proposed to learn compressed pathological representation from the gigapixel WSIs. To reconstruct the genomic features from WSIs, we introduce LD-VAE to learn the genomic posterior by which we can further learn function-specific posteriors for generating genomic features with diverse biological functions. Furthermore, we use a product-of-experts to obtain the joint latent distribution and introduce an alignment loss to improve modality consistency and enhance the estimation of the joint distribution. Finally, the co-attention module is applied to guide the selection of survival-associated instances, which is then combined with the reconstructed genomic features for survival prediction.

\subsection{Problem Formulation.}
\label{sec:method_problemformulation}
We aim to predict time-to-event results for survival analysis, where the event outcome is not always observed (\emph{e.g.} right-censored data); these predictions are obtained to estimate the probability of an event occurring before a specified time. For the $k$-$th$ patient, we can model the survival function $f^{(k)}_{surv}(T \geq t, I^{(k)})$ and hazard function $f^{(k)}_{hazard}(T=t|T \geq t, I^{(k)})$ given relative information $I^{(k)} = (P^{(k)}, G^{(k)}, c^{(k)}, t^{(k)})$, where $P^{(k)}$ represents the set of WSIs, $G^{(k)}$ denotes the genomic profiles, $c^{(k)} \in \{ 0,1 \}$ indicates the right uncensorship status, and $t^{(k)} \in \mathbb{R}^+$ is the overall survival time (in months). 
We learn the representation $f(I^{(k)})$ for the computation of the survival loss $\mathcal{L}_{surv}(\{ f(I^{(k)}), t^{(k)}, c^{(k)} \}_{k=1}^{N_D})$, where $N_D$ is the number of samples in the training sets. Specifically, we adopt the negative log-likelihood (NLL) loss~\cite{zadeh2020bias} as the objective function:  
\begin{equation}
\begin{aligned}
\mathcal{L}_{surv} &= -\sum_{k=1}^{N_D} c^{(k)} \log (f_{surv}^{(k)} (t|f(I^{(k)}))) \\ 
&+ (1-c^{(k)}) \log (f_{surv}^{(k)}(t-1|f(I^{(k)}))) \\
&+ (1-c^{(k)}) \log (f_{hazard}^{(k)}(t|f(I^{(k)})))
\label{eq:loss_surv}
\end{aligned}
\end{equation}

\subsection{Data Construction}
\label{sec:method_dataconstruction}

\subsubsection{Pathological Features Preparation}
Given an input WSI $P^{(k)}$ for the $k$-$th$ patient, we first segment the tissue regions from it and then crop it into a set of non-overlapping patches at 10$\times$ magnification, which can be denoted as $\{p^{(k)}_j\}_{j=1}^{M}$, where $M$ represents the total number of patches.
Next, a pre-trained feature extractor $\varphi_{p}(\cdot)$ (we use CTransPath~\cite{wang2021transpath, wang2022transformer} in this work) following with a fully connected layer is employed to extract the pathological features $Y^{(k)}$ as follows:
\begin{equation}
Y^{(k)}=\{\varphi_{p}(p^{(k)}_j)\}_{j=1}^{M}=\{ y^{(k)}_1, y^{(k)}_2,..., y^{(k)}_{M} \}
\end{equation}

\subsubsection{Genomic Features Preparation}
Genomic data are typically quantified as $1 \times 1$ measurements. Following previous works \cite{chen2021multimodal, xu2023multimodal}, the genomic data can be grouped into $N=6$ functional categories: 1)Tumor Suppression, 2) Oncogenesis, 3) Protein Kinases, 4) Cellular Differentiation, 5) Transcription, and 6) Cytokines and Growth. 
We denote the grouped genomic data for the $k$-$th$ patient as $\{g^{(k)}_i\}_{i=1}^{N}$. Then, a fully connected layer $\varphi_{g}(\cdot)$ introduced in~\cite{chen2021multimodal} is applied to extract the genomic features for the $k$-$th$ patient as:
\begin{equation}
X^{(k)}=\{\varphi_{g}(g^{(k)}_i)\}_{i=1}^{N}=\{ x^{(k)}_1, x^{(k)}_2,..., x^{(k)}_{N} \}
\end{equation}

\subsection{Conditional VAE}
The conditional variational autoencoder (CVAE) \cite{sohn2015learning} performs the variational inference with condition from the prior latent distribution. In this study, our goal is to generate target genomic features $X$ with pathological features $Y$.
Mathematically, the objective function of the evidence lower bound (ELBO) in CVAE under our settings can be formulated as (more details can be found in the \emph{Supplementary Material}): 
\begin{equation}
\begin{aligned}
\mathcal{L}_{ELBO}=-\mathbb{E}_{q_{\phi}(z|X,Y)}log{p_{\theta}(X|z,Y)} \\
+{\beta}KL[q_{\phi}(z|X,Y), p(z)]  
\label{eq:elbo_cvae}
\end{aligned}
\end{equation}
where the likelihood $p_{\theta}(X|z,Y)$ is parameterized by deep neural network (decoder) with $\theta$, $q(z|X,Y)$ refers to the joint posterior of pathology and genomics, and $p(z)$ represents a prior distribution.

\subsection{Pathological VIB Transformer}
\label{sec:method_vibtrans}
The CVAE is conditioned on pathological features, however, WSIs are consisted of huge number of instances that are hard for representation. 
Accordingly, we propose the Pathological Variational Information Bottleneck Transformer (VIB-Trans) that can effectively generate the bag-level representation. 
Specifically, following the goal of Information Bottleneck (IB)~\cite{tishby2000information}, we aim to learn $z_Y$ compressed from the pathological features $Y$ while maximally conserving useful information for the survival prediction target $T$. Then, we use the Variational Information Bottleneck (VIB)~\cite{alemi2016deep} to approximate the computation of IB by using variational inference with the following objective function as (more details can be found in the \emph{Supplementary Material}):
\begin{equation}
\begin{aligned}
\mathcal{L}_{VIB\text{-}Trans} = \mathbb{E}_{z_Y \sim p(z_Y|Y)}[\mathcal{L}_{surv(z_Y,t,c)}] \\ 
+ \beta KL(p(z_Y|Y), r(z_Y))
\label{eq:loss_vibtrans}
\end{aligned}
\end{equation}
where the $\mathcal{L}_{surv(z_Y,t,c)}$ is the survival prediction loss in \cref{eq:loss_surv}. $r(z_Y)$ and $p(z_Y|Y)$ correspond to the variational approximation of $p(z_Y)$ and posterior distribution over $z_Y$, respectively. 

\begin{figure}
    \centering
    \includegraphics[width=1.0\linewidth]{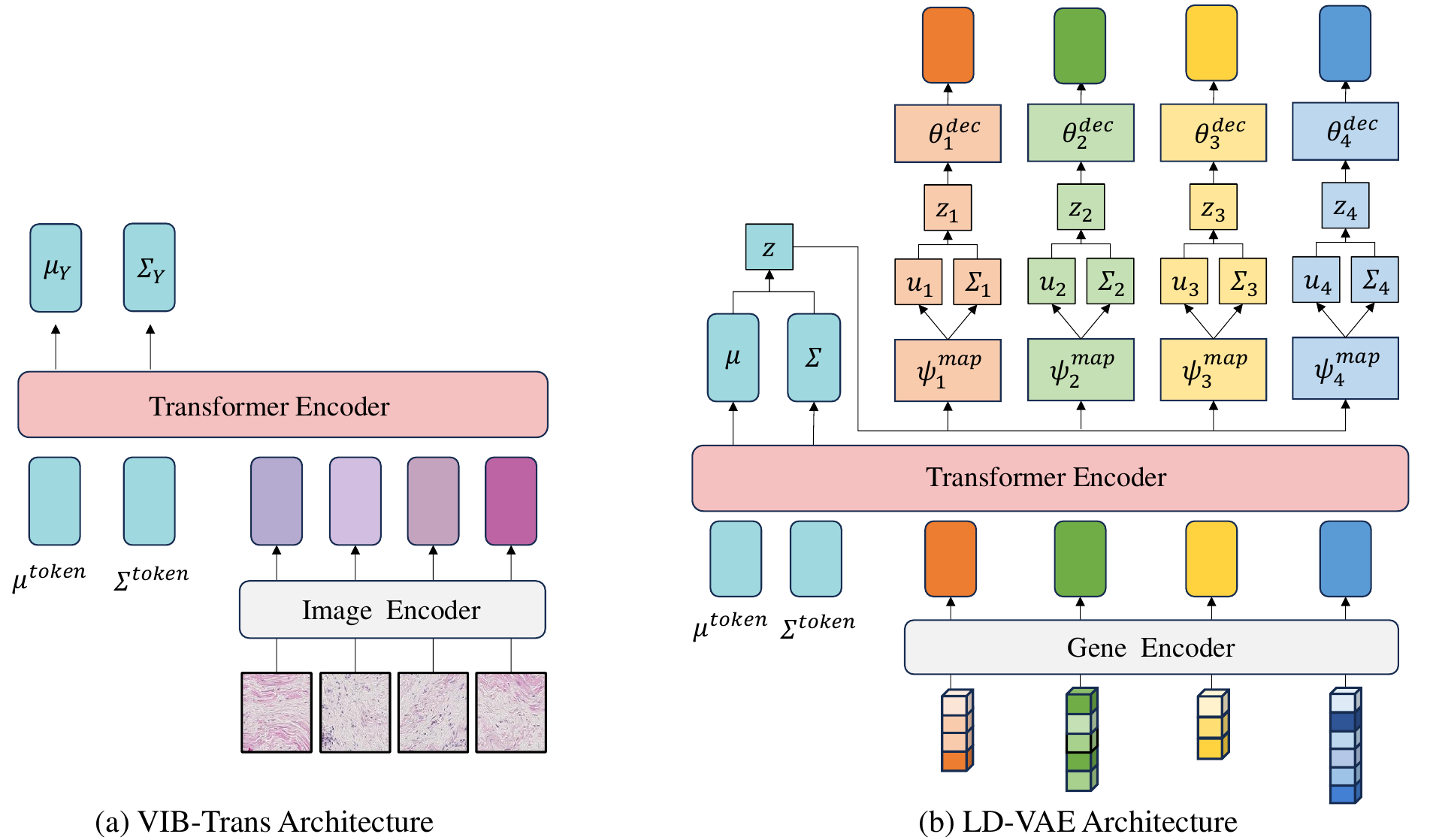}
    \caption{The architectures of (a) VIB-Trans and (b) LD-VAE. Both VIB-Trans and LD-VAE obtain posterior parameters by employing their respective transformer encoder with two learnable tokens, \emph{i.e.,} $\mu^{token}$ and $\Sigma^{token}$. Additionally, LD-VAE further learns the function-specific posteriors from the genomic posterior $\mu$ and $\Sigma$.}
    \label{fig:architecture}
\end{figure}

In practice, the posterior $p(z|Y)$ can be approximated by an encoder in the form of $p(z_Y|Y)=\mathcal{N}(z_Y|f_{e}^{\mu_Y}(Y), f_{e}^{\Sigma_Y}(Y))$, where $f_e$ is an MLP. 
However, the traditional MLP can not capture the special correlations among different instances. Consequently, we add the transformer encoder to VIB denoted as VIB-Trans (see \cref{fig:architecture}(a) for the architecture of VIB-Trans).
Specifically, VIB-Trans takes the instance-level features $Y=\{y_1, y_2, ..., y_{M}\}$ as input, and outputs the distribution parameters $\mu_Y$ and $\Sigma_Y$ in VIB. To embed arbitrary sizes of pathological features into a shared latent space, we apply the pooling operation to obtain the bag-level representation $\mu_Y$ and $\Sigma_Y$. 
Inspired by the $\textbf{[class]}$ token, which has been widely used in other domains for information aggregation~\cite{dosovitskiy2020image}, we also set two learnable tokens $\mu_Y^{token}$ and $\Sigma_Y^{token}$ to learn the global information of pathological features for the construction of the posterior distribution $p(z_Y|Y)$. 
To tackle the computation burden in traditional self-attention \cite{vaswani2017attention}, we employ the Nystrom attention \cite{xiong2021nystromformer, shao2021transmil, zhou2023cross} for approximation.
In summary, VIB-Trans can describe the global information of WSIs with the compressed representation $z_{Y}$.



\subsection{Conditional Latent Differentiation VAE}
\label{sec:method_jcld-vae}

\subsubsection{Latent Differentiation VAE}
\label{sec:method_ldvae}
In general, genomic features extracted from different functional groups usually have diverse statistical properties \cite{liberzon2015molecular}, making it challenging to generate all genomic features in a unified generative model. Therefore, we propose the Latent Differentiation Variational AutoEncoder (LD-VAE) for the reconstruction of genomic features $X=\{ x_1, x_2,...,x_N \}$ with $N$ functional categories. 
Specifically, we assume that each genomic feature in $X$ is conditionally independent given a latent variable  $z_X$. Then, the objective of VAE can be optimized with the ELBO, and the loss function can be defined as follows:
\begin{equation}
\begin{aligned}
\mathcal{L}_{ELBO} = - \mathbb{E}_{q_{\phi}(z_X|X)}[\sum^{N}_{i=1} \log p_{\theta}(x_i|z_X)] \\
+ \beta KL[q_{\phi}(z_X|X), p(z_X)]
\label{eq:elbo_ldvae1}
\end{aligned}
\end{equation}
However, it is difficult to generate different functional genomic features $x_i$ directly from the genomic posterior $p(z|x_1, x_2,...,x_N)$ as solely the genomic posterior will affect the diversity of the generated genomic features~\cite{wu2018multimodal}. To address this, we introduce the function-specific posteriors $p(z_{i}|X)$ by applying a latent differentiation process that transforms the genomic posterior into function-specific posteriors. Therefore, we can establish a shared latent space on all genomic features as well as refining the function-specific posteriors for each genomic category.

Specifically, we assign a unique latent variable $z_{i}$ to each $x_i$ and assume that $x_i$, $z_{i}$ and $z_X$ satisfy the Markov chain $x_i \leftrightarrow z_{i} \leftrightarrow z_{X}$.
Thus, we can reformulate $\log p(x_i|z_X)$ in \cref{eq:elbo_ldvae1} as follows (more details are provided in the \emph{Supplementary Material}):
\begin{equation}
\log p(x_i|z_X) = \log p(x_i, z_{i} | z_X) 
= \log p(x_i | z_{i}) p(z_{i}|z_X)
\end{equation}
Then the loss function of LD-VAE can be derived as:
\begin{equation}
\begin{aligned}
\mathcal{L}_{ELBO} = - \mathbb{E}_{q_{\phi}(z_X|X)}[\sum^{N}_{i=1} \log p_{\theta}(x_i|z_{i}) p(z_{i}|z_X)] \\
+ \beta (\sum^{N}_{i=1} KL[q_{\phi}(z_{i}|X), p(z_{i})] + KL[q_{\phi}(z_X|X), p(z_X)])
\label{eq:elbo_ldvae2}
\end{aligned}
\end{equation}
Here, $p(z_{i}|z_X)$ can be approximated in the form of $p(z_{i}|z_X)=\mathcal{N}(z_{i}|\psi_{i}^{\mu}(z_X), \psi_{i}^{\Sigma}(z_X))$ where $\psi_i$ serves as an MLP mapper. In practice, each function-specific posterior should also approximate prior distribution $r(z_{i})$, which can be accomplished by minimizing the Kullback-Leibler divergence.
The detailed architecture of LD-VAE has been shown in \cref{fig:architecture}(b), and we provide more details in the \emph{Supplementary Material}.

\subsubsection{Conditional Latent Differentiation VAE}
The Conditional Latent Differentiation VAE (LD-CVAE) adds the condition of pathological features $Y$ into LD-VAE for the reconstruction of different functional genomic features $X$, and we can derive the loss function of ELBO in LD-CVAE by combining \cref{eq:elbo_cvae} and \cref{eq:elbo_ldvae2} as follows:
\begin{equation}
\begin{aligned}
\mathcal{L}_{ELBO} = - \mathbb{E}_{q_{\phi}(z|X, Y)}[\sum^{N}_{i=1} \log p_{\theta}(x_i|z_{i}, Y) p(z_{i}|z)] \\
+ \beta (\sum^{N}_{i=1} KL[q_{\phi}(z_{i}|X, Y), p(z_{i})] + KL[q_{\phi}(z|X, Y), p(z)])
\label{eq:elbo_ldvae3}
\end{aligned}
\end{equation}
where $q_{\theta}(z|X,Y)$ represents the joint posterior of the pathological and genomic features. For the purpose of computing $\log p_{\theta} (x_i|z_{i}, Y)$, we additionally incorporate the latent representation $z_Y$ into the reconstruction network $\theta_i$. 

\subsubsection{Joint Pathology-Genomics Distribution Learning}
\label{sec:method_jointdistributionlearning}
One critical problem for optimizing \cref{eq:elbo_ldvae3} is to estimate the joint posterior $q(z|X,Y)$. According to the previous work \cite{wu2018multimodal}, a product-of-experts (PoE) can be applied to approximate the joint posterior $q(z|X,Y)$ by the combination of the marginal posteriors $\tilde{q}(z|X)$ and $\tilde{q}(z|Y)$, with the following formulation (derivation can be found in the \emph{Supplementary Material}):
\begin{equation}
p(z|X,Y) \approx \tilde{q}(z|X) \tilde{q}(z|Y) p(z) \equiv q(z|X, Y)
\end{equation}
Here, we approximate $p(z|X) \equiv \tilde{q}(z|X) p(z)$, $p(z|Y) \equiv \tilde{q}(z|Y) p(z)$ where $\tilde{q}(\cdot)$ is the underlying inference network.
Specifically, $\tilde{q}(z|X)$ and $\tilde{q}(z|Y)$ can be represented as $\tilde{q}(z|X)=\mathcal{N}(z|\mu_X,\Sigma_X)$ and $\tilde{q}(z|Y)=\mathcal{N}(z|\mu_Y,\Sigma_Y)$, which can be derived from the variational genomic posterior of genomic features (shown in \cref{eq:elbo_ldvae2}) and variational posterior of pathological features (shown in \cref{eq:loss_vibtrans}). In practice, $p(z)$ is a spherical Gaussian distribution $\mathcal{N}(z|x_0,y_0)$. Since the product of Gaussian still follows the Gaussian distribution, we can derive the joint posterior in the closed form as: $q(z|X,Y)=\mathcal{N}(\mu, \Sigma)$, where $\Sigma = ( \Sigma_0^{-1}+\sum\nolimits_{i \in \{ X,Y \}} \Sigma_{i}^{-1})^{-1}, \mu = (\mu_0 \Sigma_0^{-1} + \sum\nolimits_{i \in \{X,Y\}} \mu_i \Sigma_i^{-1} \Sigma^{-1})$. 
It worth noting that the employment of PoE allows us to learn the joint posterior using complete multimodal data during training and approximate the joint posterior distribution during testing, even in the absence of genomic data. 

To enhance the consistency between pathological features and genomic features, we propose an alignment loss that optimizes the Wasserstein distance~\cite{givens1984class} between the marginal distributions $\tilde{q}(z|X)=\mathcal{N}(z|\mu_X,\Sigma_X)$ and $\tilde{q}(z|Y)=\mathcal{N}(z|\mu_Y,\Sigma_Y)$, which can be denoted as follows:
\begin{equation}
\mathcal{L}_{Align}=||\mu_X - \mu_Y||^2 + \textbf{Tr}(\Sigma_X + \Sigma_Y - 2{(\Sigma_Y^{1/2} \Sigma_X \Sigma_Y^{1/2})}^{1/2} )
\label{eq:loss_align}
\end{equation}
Since the covariance matrices are diagonal, it can be further simplified to:
\begin{equation}
\mathcal{L}_{Align} = ||\mu_X - \mu_Y||^2 + ||\Sigma_X - \Sigma_Y||^2
\label{eq: align}
\end{equation}

\subsubsection{The Objective Function of LD-CVAE} By the combination of \cref{eq:elbo_ldvae3} and \cref{eq: align}, the final objective function of LD-CVAE can be formulated as follows: 
\begin{equation}
\mathcal{L}_{LD\text{-}CVAE}=\mathcal{L}_{ELBO} + \alpha \mathcal{L}_{Align}
\label{eq:ldcvae}
\end{equation}
where $\alpha$ controls the impact of $\mathcal{L}_{Align}$.

\subsection{Multimodal Fusion and Prediction}
After obtaining the pathological features $Y$ and the reconstruction genomic features $\tilde{X}$ from the LD-CVAE, we combine them to obtain the fused representation.
Specifically, we apply the Genomic-Guided Co-Attention Layer reported in \cite{chen2021multimodal} to model pairwise interactions between the instance-level pathological features $Y$ and their corresponding genomic features $\tilde{X}$, which can be formulated as 
\begin{equation}
CoAttn_{\tilde{X} \to Y}(\tilde{X},Y)=\text{softmax}(\frac{\textbf{W}_q \tilde{X} (\textbf{W}_k Y)^{\top}}{\sqrt{d_k}}) \textbf{W}_v Y 
\end{equation}
We further apply two set-based MIL Transformers $\mathcal{T}_{P}$,$\mathcal{T}_{G}$ \cite{chen2021multimodal} with the global attention pooling function $\mathcal{F}_{\text{attnpool}(\cdot)}$ to construct the bag-level pathological and genomic features by aggregating feature embeddings of $\hat{Y}$ and $\tilde{X}$. The two embeddings are concatenated and then fed into the fully connected layer to obtain the final feature embeddings for survival prediction.

\paragraph{Overall Loss.} The final loss consists of three terms: the survival loss $\mathcal{L}_{surv}$, VIB-Trans loss $\mathcal{L}_{VIB\text{-}Trans}$ and LD-CVAE loss $\mathcal{L}_{LD\text{-}CVAE}$, which can be formulated as:
\begin{equation}
\mathcal{L} = \mathcal{L}_{surv} + \mathcal{L}_{VIB\text{-}Trans} + \mathcal{L}_{LD\text{-}CVAE} 
\end{equation}
\section{Experiments}
\label{sec:experiments}

\begin{table*}[htbp]
\setlength{\arrayrulewidth}{0.1mm}  
\tiny 
\centering
\caption{Comparisons of C-index (mean ± std) with SOTA methods over five cancer datasets. g. and h. refer to solely relies on image and genomic data, respectively. The best results and the second-best results are highlighted with \textbf{bold} and in \underline{underline}, respectively.}
\label{table:table1}
\resizebox{\linewidth}{!}{
\begin{tabular}{c|c|cccccc}
\hline
\hline
\multirow{2}{*}{\textbf{Model}}  & \multirow{2}{*}{\textbf{Modality}} & \textbf{BLCA} & \textbf{BRCA} & \textbf{GBMLGG} & \textbf{LUAD} & \textbf{UCEC} & \multirow{2}{*}{\textbf{Overall}} \\
 &  & (N=373) & (N=957) & (N=571) & (N=452) & (N=480) &  \\
\hline
\hline
MLP & g. & 0.613   \texttt{\textpm} 0.019  & 0.587 \texttt{\textpm} 0.033  & 0.809 
\texttt{\textpm} 0.029  & 0.617  \texttt{\textpm} 0.026  & 0.657  \texttt{\textpm} 0.036  & 0.657 \\
SNN & g. & 0.619  \texttt{\textpm} 0.023  & 0.596  \texttt{\textpm} 0.027  & 0.805 \texttt{\textpm} 0.030   & 0.625  \texttt{\textpm} 0.019  & 0.651  \texttt{\textpm} 0.018  & 0.659  \\
SNNTrans & g. & 0.627 \texttt{\textpm} 0.019 & 0.618  \texttt{\textpm} 0.018 & 0.816  \texttt{\textpm} 0.037  & 0.631  \texttt{\textpm} 0.023  & 0.641 \texttt{\textpm} 0.026 & 0.667 \\
\hline
ABMIL & h. & 0.622 \texttt{\textpm} 0.051  & 0.614 \texttt{\textpm} 0.037 & 0.786  \texttt{\textpm} 0.028  & 0.596  \texttt{\textpm} 0.067  & 0.649  \texttt{\textpm} 0.028  & 0.652  \\
CLAM-SB & h. & 0.613  \texttt{\textpm} 0.058  & 0.607  \texttt{\textpm} 0.016 & 0.792  \texttt{\textpm} 0.032  & 0.590  \texttt{\textpm} 0.076  & 0.650 \texttt{\textpm} 0.066   & 0.650    \\
CLAM-MB & h. & 0.621  \texttt{\textpm} 0.065  & 0.617 \texttt{\textpm} 0.012  & 0.799  \texttt{\textpm} 0.028  & 0.596  \texttt{\textpm} 0.075  & 0.657  \texttt{\textpm} 0.028   & 0.658   \\
TransMIL & h. & 0.625  \texttt{\textpm} 0.050  & 0.614  \texttt{\textpm} 0.037 & 0.803  \texttt{\textpm} 0.037  & 0.593  \texttt{\textpm} 0.063  & 0.665  \texttt{\textpm} 0.019  & 0.660  \\
\hline
Porpoise & g.+h. & 0.661  \texttt{\textpm} 0.034  & 0.644  \texttt{\textpm} 0.010 
 & 0.829  \texttt{\textpm} 0.033  & 0.646  \texttt{\textpm} 0.025  & 0.660 \texttt{\textpm} 0.106 & 0.694   \\
MCAT & g.+h. & 0.660  \texttt{\textpm} 0.046  & 0.652  \texttt{\textpm} 0.019  & 0.833  \texttt{\textpm} 0.029  & 0.642  \texttt{\textpm} 0.054  & 0.686  \texttt{\textpm} 0.042  & 0.695  \\
MOTCat & g.+h. & 0.667  \texttt{\textpm} 0.038  & 0.659  \texttt{\textpm} 0.034  & 0.835  \texttt{\textpm} 0.037  & 0.656  \texttt{\textpm} 0.021  & 0.683  \texttt{\textpm} 0.060  & 0.700  \\
CMTA & g.+h. & \underline{0.671  \texttt{\textpm} 0.039}  & 0.648  \texttt{\textpm} 0.048  & \underline{0.846  \texttt{\textpm} 0.041}  & 0.666  \texttt{\textpm} 0.025  & 0.699  \texttt{\textpm} 0.035  & 0.706  \\
SurvPath & g.+h. & 0.663  \texttt{\textpm} 0.026  & \underline{0.661 \texttt{\textpm} 0.024} & 0.839  \texttt{\textpm} 0.039  & \underline{0.671  \texttt{\textpm} 0.036}  & \textbf{0.709  \texttt{\textpm} 0.039}  & \underline{0.709}  \\
PIBD & g.+h. & 0.634  \texttt{\textpm} 0.029  & 0.636  \texttt{\textpm} 0.028  & 0.817  \texttt{\textpm} 0.038  & 0.628  \texttt{\textpm} 0.063  & 0.687  \texttt{\textpm} 0.072  & 0.680  \\
\rowcolor{gray!20}
Ours & g.+h. & \textbf{0.686  \texttt{\textpm}
 0.035}  & \textbf{0.680  \texttt{\textpm}
 0.030}  & \textbf{0.849  \texttt{\textpm}
 0.017}  & \textbf{0.676  \texttt{\textpm}
 0.015}  & \underline{0.703 \texttt{\textpm} 0.069}  & \textbf{0.719}  \\
\hline
\hline
\end{tabular}
}
\end{table*}

\begin{table*}[htbp]
\setlength{\arrayrulewidth}{0.1mm}  
\tiny 
\centering
\caption{Comparisons of C-index (mean ± std) with methods addressing missing modality over five cancer datasets. The best results and the second-best results are highlighted in \textbf{bold} and in \underline{underline}, respectively}
\label{table:table2}
\setlength{\tabcolsep}{8pt}
\resizebox{\linewidth}{!}{
\begin{tabular}{ccccccc}
\hline
\hline
\textbf{Model}   & \textbf{BLCA} & \textbf{BRCA} & \textbf{GBMLGG} & \textbf{LUAD} & \textbf{UCEC} & \textbf{Overall} \\
\hline
\hline
VAE  & 0.622  \texttt{\textpm} 0.010  & 0.629  \texttt{\textpm} 0.020  & 0.805  \texttt{\textpm} 0.032  & 0.598  \texttt{\textpm} 0.029  & 0.660  \texttt{\textpm} 0.029  & 0.663  \\
GAN  & 0.621  \texttt{\textpm} 0.018  & 0.621  \texttt{\textpm} 0.027  & 0.793  \texttt{\textpm} 0.045  & 0.608  \texttt{\textpm} 0.028  & 0.663  \texttt{\textpm} 0.036  & 0.661  \\
MVAE  & 0.629  \texttt{\textpm} 0.009  & 0.619  \texttt{\textpm} 0.027  & 0.790  \texttt{\textpm} 0.018  & 0.610  \texttt{\textpm} 0.030  &0.661 \texttt{\textpm} 0.024 & 0.662 \\
SMIL  & 0.627  \texttt{\textpm} 0.015  & 0.610  \texttt{\textpm} 0.010  & 0.807  \texttt{\textpm} 0.012  & 0.608  \texttt{\textpm} 0.035  & \underline{0.678 \texttt{\textpm} 0.016} & 0.666 \\
ShaSpec  & \underline{0.630  \texttt{\textpm} 0.031}  & \underline{0.626  \texttt{\textpm} 0.027}  & 0.810  \texttt{\textpm} 0.024  & \underline{0.613  \texttt{\textpm} 0.036}  & 0.672
  \texttt{\textpm} 0.037  & \underline{0.670} \\
Transformer  & 0.629 \texttt{\textpm} 0.022  & 0.621  \texttt{\textpm} 0.046  & \underline{0.814  \texttt{\textpm} 0.016}  & 0.610  \texttt{\textpm} 0.020  & 0.673  \texttt{\textpm} 0.012  & 0.669  \\
\rowcolor{gray!20}
Ours  & \textbf{0.649  \texttt{\textpm} 0.040}  & \textbf{0.641  \texttt{\textpm} 0.012}  & \textbf{0.821  \texttt{\textpm} 0.021}  & \textbf{0.628  \texttt{\textpm} 0.008}  & \textbf{0.681  \texttt{\textpm} 0.044}  & \textbf{0.684} \\
\hline
\hline
\end{tabular}
}
\vspace{-1mm}
\end{table*}

\renewcommand{\thefootnote}{\arabic{footnote}}

\subsection{Datasets and Settings}
\paragraph{Datasets and Evaluation.} To validate the effectiveness of our proposed method, we use five different cancer datasets from The Cancer Genome Atlas (TCGA) \footnote{\url{https://portal.gdc.cancer.gov/}}: Bladder Urothelial Carcinoma (BLCA), Breast Invasive Carcinoma (BRCA), Glioblastoma and Lower Grade Glioma (GBMLGG), Lung Adenocarcinoma (LUAD), and Uterine Corpus Endometrial Carcinoma (UCEC). We employ 5-fold cross-validation for each dataset. The performance of our model is assessed by concordance index (C-index) \cite{harrell1996multivariable}. Additionally, we employ the Kaplan-Meier (KM) survival curves \cite{kaplan1958nonparametric} and Log-rank test~\cite{mantel1966evaluation} to test the patient stratification performance among different methods . 

\paragraph{Implementation.}
For each WSI, we employ a pre-trained Swin Transformer encoder (CTransPath) \cite{wang2021transpath, wang2022transformer} as the pathological encoder $\varphi_{p}$. For genomic data, we use the SNN \cite{klambauer2017self} as the  encoder $\varphi_{g}$. During the training process, we empirically set the parameter $\alpha$ in \cref{eq:ldcvae} as 0.1. The hyperparameter $\beta$ for Kullback-Leibler divergence is annealed using a cosine schedule, and thus forming a valid lower bound on the evidence. More discussions for the parameter settings can be found in the \emph{Supplementary Material}. We adopt Adam optimizer with the initial learning rate of 2$\times$10$^{-4}$ and weight decay of 1$\times$10$^{-5}$. Following the setting of \cite{chen2021multimodal}, we use the batch size of 1 for model training. 

\begin{table*}[htb]
\setlength{\arrayrulewidth}{0.1mm}  
\tiny 
\centering
\caption{Ablation study under both complete modality and missing modality scenarios.}
\label{table:table_ablation}
\vspace{-1mm}
\resizebox{\linewidth}{!}{
\begin{tabular}{c|c|cccccc}
\hline
\hline
\textbf{Variants} & \textbf{Missing}  & \textbf{BLCA} & \textbf{BRCA} & \textbf{GBMLGG} & \textbf{LUAD} & \textbf{UCEC} & \textbf{Overall} \\
\hline
\hline
\multirow{2}{*}{w/o VIB-Trans} & & 0.651 \texttt{\textpm}	0.035 & 0.685 \texttt{\textpm}	0.020 &	0.833 \texttt{\textpm}	0.058 &	0.646 \texttt{\textpm}	0.048 &	0.691 \texttt{\textpm}	0.053 &	0.701 \\
& \checkmark & 0.616 \texttt{\textpm}	0.024 &	0.619 \texttt{\textpm}	0.019 &	0.820 \texttt{\textpm}	0.040 &	0.623 \texttt{\textpm}	0.035 &	0.647 \texttt{\textpm}	0.037 &	0.665 \\
\hline
\multirow{2}{*}{w/o LD-VAE}
  && 0.674 \texttt{\textpm}	0.017 &	0.671 \texttt{\textpm}	0.024 & 0.835 \texttt{\textpm}	0.027 &	0.668 \texttt{\textpm}	0.024 &	0.688 \texttt{\textpm}	0.025 &	0.707 \\
 & \checkmark & 0.641 \texttt{\textpm}	0.035 &	0.634 \texttt{\textpm}	0.048 &	0.814 \texttt{\textpm}	0.019 &	0.616 \texttt{\textpm}	0.012 &	0.672 \texttt{\textpm}	0.056 &	0.675 \\
\hline
\multirow{2}{*}{w/o Alignment Loss}  & & 0.663 \texttt{\textpm}	0.024 &	0.676 \texttt{\textpm}	0.040 &	0.828 \texttt{\textpm}	0.033 &	0.659 \texttt{\textpm}	0.017 &	0.689 \texttt{\textpm}	0.024 &	0.703 \\
& \checkmark  & 0.643 \texttt{\textpm}	0.031 &	0.629 \texttt{\textpm}	0.026 &	0.783 \texttt{\textpm}	0.019 &	0.621 \texttt{\textpm}	0.021 &	0.670 \texttt{\textpm}	0.036 &	0.669 \\
\hline
\multirow{2}{*}{Ours}  & & {0.686  \texttt{\textpm} 0.035}  & {0.680  \texttt{\textpm} 0.030}  & {0.849  \texttt{\textpm} 0.017}  & {0.676  \texttt{\textpm}0.015}  & {0.703  \texttt{\textpm} 0.069}  & {0.719} \\
 & \checkmark  & 0.649  \texttt{\textpm} 0.040  & 0.641  \texttt{\textpm} 0.012  & 0.821  \texttt{\textpm} 0.021  & 0.628  \texttt{\textpm} 0.008  & 0.681  \texttt{\textpm} 0.044  & 0.684 \\
\hline
\hline
\end{tabular}
}
\vspace{-1mm}
\end{table*}

\begin{figure*}
    \vspace{-2mm}
    \centering
    \includegraphics[width=1.0\linewidth]{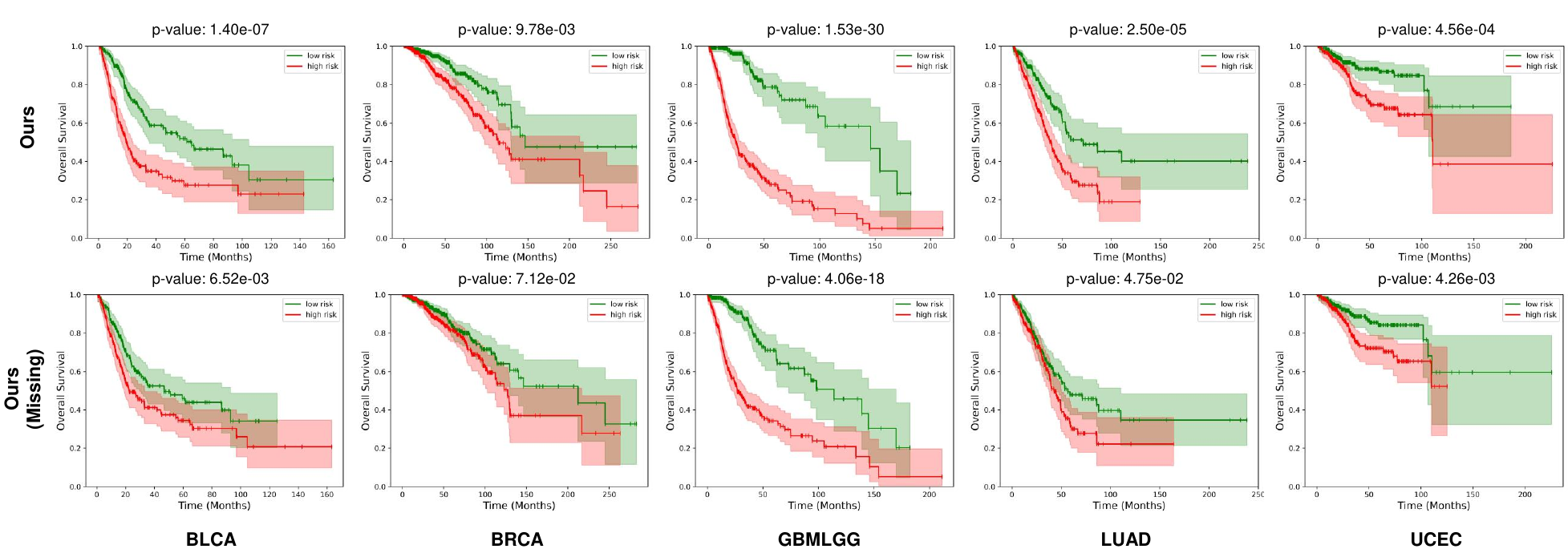}
    \caption{Kaplan-Meier Analysis of predicted high-risk (red) and low-risk (green) groups on five cancer datasets under both complete modality (top) and missing modality (bottom) scenarios. Shaded areas refer to the confidence intervals.}
    \label{fig:exp_curves_ours}
    \vspace{-3mm}
\end{figure*}

\begin{figure}
    \centering
    \includegraphics[width=1.0\linewidth]{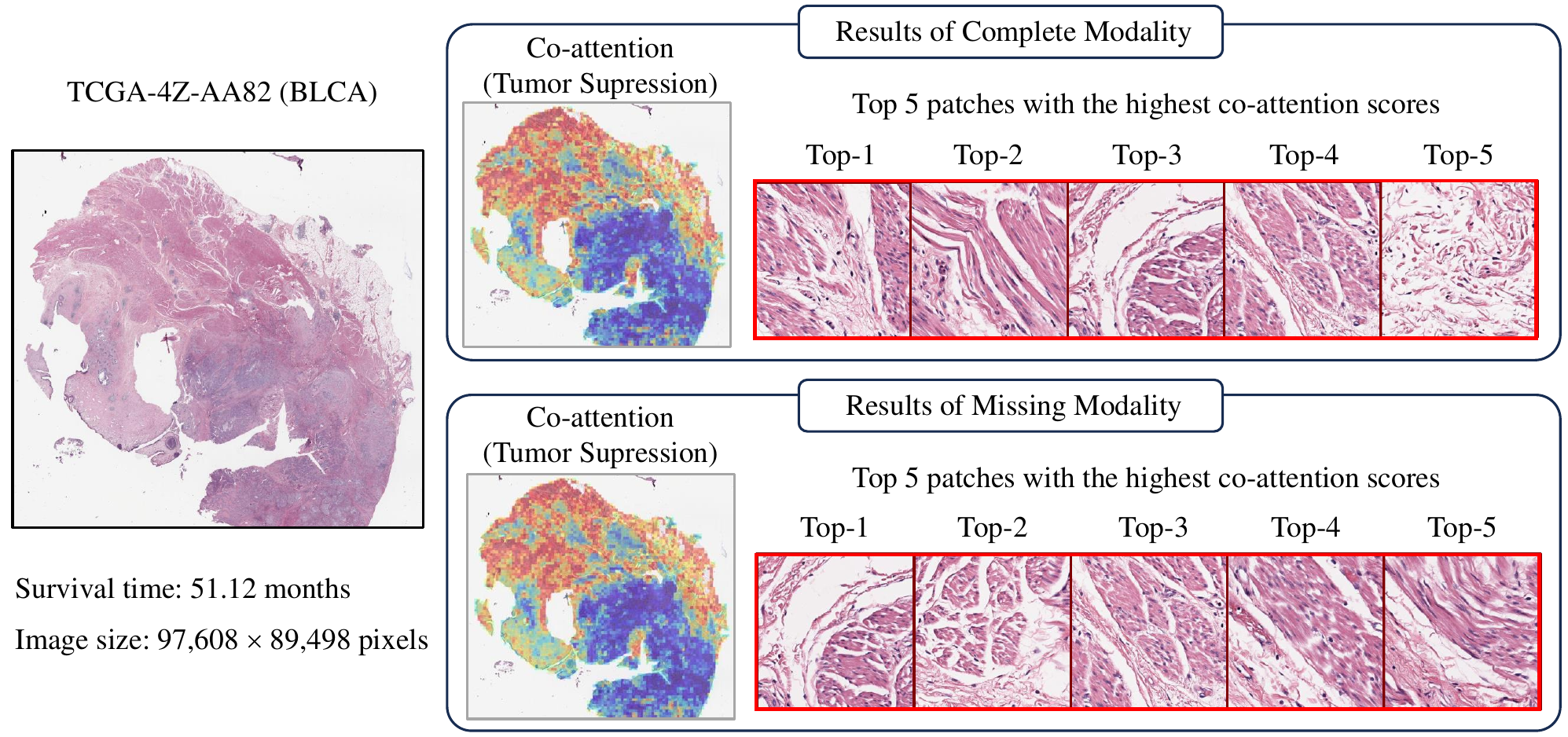}
    \caption{Comparison of the co-attention weights calculated from the genuine (top) and generated (bottom) genomic features.}
    \label{fig:exp_visulization}
    \vspace{-4mm}
\end{figure}

\subsection{Comparisons with State-of-the-Arts}
To demonstrate the effectiveness of our model on complete multimodal data, we compare it with state-of-the-art (SOTA) methods for both unimodal and multimodal survival analysis. For \textbf{unimodal methods}, we consider genomic-based approaches such as MLP~\cite{haykin1998neural}, SNN~\cite{klambauer2017self} and SNNTrans~\cite{klambauer2017self, shao2021transmil}, along with histopathological baselines like ABMIL~\cite{ilse2018attention}, CLAM~\cite{lu2021data},  TransMIL~\cite{shao2021transmil}. For \textbf{multimodal methods}, we compare our model with  Porpoise~\cite{chen2022pan}, MCAT~\cite{chen2021multimodal}, MOTCat~\cite{xu2023multimodal}, CMTA~\cite{zhou2023cross}, SurvPath~\cite{jaume2024modeling}, and PIBD~\cite{zhang2024prototypical}. In addition, to further verify the effectiveness of our model with missing genomic data, we compare it with the following studies that can \textbf{address missing modality data for multimodal learning}: (a) VAEs~\cite{alemi2016deep} and GANs~\cite{goodfellow2014generative} are both generative models and we can use them to generate the missing genomic data; (b) MVAE~\cite{wu2018multimodal} is a multimodal generative model that can tackle missing modality in the testing stage. (c) SMIL~\cite{ma2021smil} is proposed to handle missing modality in multimodal learning by leveraging Bayesian meta-learning. (d) ShaSpec~\cite{wang2023multi} fuses shared and specific features to address the missing modality data. (e) Robust Multimodal Transformer~\cite{ma2022multimodal} is presented to improve the robustness of multimodal Transformer via multi-task learning.

\vspace{-3mm}
\paragraph{Compared with Unimodal Models.} From the results in \cref{table:table1}, we can observe that the multimodal methods consistently outperform unimodal approaches, demonstrating the advantage of the integrative analysis of genomics and histology for survival analysis. Note that our approach outperforms its unimodal counterparts across all datasets. Specifically, compared to the best model relying on genomic data, our method can improve the overall performance by 5.2\%. On the other hand, our method also surpass the best existing studies based on pathology image data by 5.9\%. 

\vspace{-3mm}
\paragraph{Compared with Multimodal Models.}
As can be seen from \cref{table:table1}, compared with all multimodal approaches, the proposed method achieves the highest performance in 4 out of 5 cancer datasets with 1.0\%-3.9\% improvement. These results clearly demonstrate the advantage of our method that can better learn genomic representation and fuse multi-modal data for survival prediction of human cancers.

\vspace{-3mm}
\paragraph{Compared with Baseline Models Addressing Missing Modality.} For our method and its competitors, we use complete data for model training and test their performance with missing genomic data. The experimental results shown in \cref{table:table2} suggest that all the studies addressing missing modality can achieve higher C-index values in comparison with the studies solely relying on the pathology image data (shown in \cref{table:table1}). On the other hand, the experimental results in \cref{table:table2} also indicate that our method can achieve higher C-index values than the comparing studies, highlighting the capability of our method in handling the missing modality of genomic data for survival prediction.

\subsection{Ablation Study}
\paragraph{Impact of VIB-Trans.}
For the variant of w/o VIB-Trans, we use TransMIL encoder as a baseline to represent WSI (additional results of other baselines are provided in the \emph{Supplementary Material}). As shown in~\cref{table:table_ablation}, compared with the TransMIL encoder, VIB-Trans improves the overall performance by 1.8\% on complete data and 1.9\%  with missing genomic data, which implies that VIB-Trans can effectively obtain the  representation of WSIs.   
\paragraph{Impact of LD-VAE.} For the variant of w/o LD-VAE, we only use the VAE architecture for the genomic posterior estimation without considering the functional difference among diverse genomic features. As shown in \cref{table:table_ablation}, 
our method surpass w/o LD-VAE by 1.2\% and 0.9\% on complete multi-modal data and the scenario with missing genomic data, respectively. These results again validate 
the advantage of our method that can generate better genomic representations for survival prediction.
\paragraph{Impact of Alignment Loss.} Additionally, we evaluate the effect of the alignment loss (shown in \cref{eq: align} in our method. As can be seen from \cref{table:table_ablation}, our method achieves higher C-index values than its variant \emph{i.e.,} w/o Alignment loss. This is because the proposed alignment loss can improve the consistency for multi-modal integration, and thus can enhance the estimation of joint posterior with marginal pathological posterior. More discussions for the ablation study can be found in the \emph{Supplementary Material}

\subsection{Patient Stratification}
Besides comparing the prognosis performance of different methods by the measurements of C-index, another important task for cancer prognosis is to stratify cancer patients into different groups with diverse clinical outcomes, by which we could develop personalized treatment for different cancer patients. In \cref{fig:exp_curves_ours}, we compare the patient stratification performance of our method under the settings of complete multi-modal data (top row of \cref{fig:exp_curves_ours}) and missing genomic data (bottom row of \cref{fig:exp_curves_ours}). As shown in \cref{fig:exp_curves_ours}, we derive similar stratification performance under these two settings across different cancer cohorts. These results further clearly demonstrate that the generated genomic feature by our method is also associated with the survival information, which in turn can help predict the clinical outcome of cancer patients. We also compare the prognosis performance of our method for patient stratification with other SOTA algorithms in the \emph{Supplementary Materials}, and our method can still achieve superior prediction results.

\subsection{Visualization Results}
In this section, we further compare the co-attention weights that are calculated from the ground truth and generated genomic feature. In \cref{fig:exp_visulization}, we visualize the co-attention weights and show the patches with \emph{top-5} scores for the genomic features with the function of \emph{Tumor Suppression}. As shown in \cref{fig:exp_visulization},  we could derive similar visualization results of co-attention score among genuine and reconstructed genomic features. In addition, their corresponding \emph{top-5} highly correlated patches also share similar visual appearance. These results again suggest the advantage of our method that can generate genomic features with close connection to the pathology images. More visualization results are provided in the \emph{Supplementary Material}.
\vspace{-2mm}
\section{Conclusion}
In this paper, we present a Conditional Latent Differentiation Variational AutoEncoder (LD-CVAE) that can reconstruct the genomic embeddings with diverse biological functions from WSIs for robust multimodal survival prediction. The experimental results on five different cancer datasets demonstrate the superiority of our method under both complete and missing modality scenarios. 

\vspace{-2mm}
\section*{Acknowledgments}
This work is supported by the National Natural Science Foundation of China (Nos.62136004, 62272226, 62402219) and the Key Research and Development Plan of Jiangsu Province (No.BE2022842).

{
    \small
    \bibliographystyle{ieeenat_fullname}
    \bibliography{main}
}

\clearpage
\setcounter{page}{1}
\maketitlesupplementary

\section{Detailed Methods}
\subsection{Survival Analysis}
\label{app:rationale}
For \emph{k-th} patient, we can model the survival and hazard functions given $I^{(k)} = (P^{(k)}, G^{(k)}, c^{(k)}, t^{(k)})$, where $P^{(k)}$ represents the set of whole slide images, $G^{(k)}$ denotes the genomic profiles, $c^{(k)} \in \{ 0,1 \}$ indicates the right uncensorship status, and $t^{(k)} \in \mathbb{R}^+$ is the overall survival time (in months). The hazard function $f^{(k)}_{hazard}(T=t|T \geq t, I^{(k)})$ measures the instantaneous risk of death at time point $k$ for the \emph{k-th} patient, which can be defined as:
\begin{equation}
f^{(k)}_{hazard} (T=t) = \lim_{\partial t \to 0} \frac{P(t \leq T \leq t+\partial t | T \geq t)}{\partial t}
\end{equation}
The survival function $f^{(k)}_{surv}(T \geq t, I^{(k})$ quantifies the probability of surviving after a specified time $t$, which can be estimated via the cumulative hazard function $f^{(k)}_{hazard}(T=t|T \geq t, I^{(k)})$ as follows:
\begin{equation}
f_{surv}(T \geq t, I^{(k)}) = \prod\nolimits_{u=1}^{t}(1-f_{hazard}(T=u|T \geq u, I^{(k)}))
\end{equation}
The most common method for estimating the hazard
function is the Cox Proportional Hazards (CoxPH) model, in which $f_{hazard}$ can be parameterized as:
\begin{equation}
\lambda (t|x) = \lambda_{0}(t) e^{\theta X}
\end{equation}
where $\lambda_{0}(t)$ represents the baseline hazard function, $\theta$ represents the vector of coefficients for the covariates.

\subsection{Conditional VAE}
The conditional variational autoencoder (CVAE) \cite{sohn2015learning} performs the variational inference with condition from the prior latent distribution. In our case, the goal is to generate target genomic features $X$ given the pathological features $Y$.
The variational lower bound of CVAE can be derived as follows:
\begin{equation}
\begin{aligned}
&\log{p_{\theta}(X|Y)} =
\int_{z}dz{q_{\phi}(z|X,Y)}{\log p_{\theta}(X|Y)} \\
&=\int_{z}dz{q_{\phi}(z|X,Y)}{\log p_{\theta}(X|Y)} \\
&=\mathbb{E}_{q_{\phi}(z|X,Y)} \log p_\theta (X|Y) \\
&=\mathbb{E}_{q_{\phi}(z|X,Y)} \log \frac{p_\theta (X,Y,z)}{p_\theta (z|X,Y) p_\theta (Y)} \\
&=\mathbb{E}_{q_{\phi}(z|X,Y)} \log \frac{q_\phi (z|X,Y)}{p_\theta (z|X,Y)} \frac{p_\theta (X,Y,z)}{q_\phi (z|X,Y) p_\theta (Y)} \\
&= KL(q_\phi (z|X,Y) || p_\theta (z|X,Y)) + \mathbb{E}_{q_{\phi}(z|X,Y)} \log \frac{p_\theta (X,Y,z)}{q_\phi (z|X,Y) p_\theta (Y)} \\
&\geq \mathbb{E}_{q_{\phi}(z|X,Y)} \log \frac{p_\theta (X,Y,z)}{q_\phi (z|X,Y) p_\theta (Y)} \\
&= \mathbb{E}_{q_{\phi}(z|X,Y)} \log \frac{p_\theta (X|z,Y) p_\theta (z|Y) p_\theta (Y)}{q_\phi (z|X,Y) p_\theta (Y)} \\
&= \mathbb{E}_{q_{\phi}(z|X,Y)} \log p_\theta (X|z,Y) - KL(q_\phi (z|X,Y) || p_\theta (Y))
\end{aligned}
\end{equation}

\subsection{Pathological VIB Transformer}
The goal of VIB \cite{alemi2016deep} is to learn a new representation $z_Y$  compressed from the original pathological information $Y$ while conserving information about the target $T$. To learn the minimal sufficient representation $z_Y$, we can formulate it as follows (for clarity, we employ $Z$ as an alternative for $z_Y$ in this section):
\begin{equation}
\arg\max_{Z} I(Z, T) - \beta I(Z, Y) 
\end{equation}
where $I(\cdot)$ represents mutual information (MI). The hyperparameter $\beta \geq 0$ controls the trade-off between compression and prediction, determining the strength of the bottleneck.
However, the computation of MI is intractable, VIB~\cite{alemi2016deep} approximates the computation of IB by using variational inference.

We firstly suppose that the joint distribution $p(Y, T, Z)$ can be calculated via:
\begin{equation}
p(Y,T,Z) = p(Z|Y,T) p(T|Y)p(Y) = p(Z|Y) p(T|Y) p(Y)
\end{equation}
where we assume that $p(Z|Y,T) = p(Z|Y)$ follows the Markov chain $T \leftrightarrow Y \leftrightarrow Z$. Then, we can reformulate the terms $I(Z, T)$ and $I(Z, Y)$ as follows:
\begin{equation}
\begin{aligned}
I(Z, T) &= \int dt \, dz \, p(z, t) \log \frac{p(z, t)}{p(z) p(t)} \\
&= \int dt \, dz \, p(z, t) \log \frac{p(t|z)}{p(z)}
\end{aligned}
\end{equation}
\begin{equation}
\begin{aligned}
I(Z, Y) &= \int dy \, dz \, p(z, y) \log \frac{p(z, y)}{p(z) p(y)} \\
&= \int dy \, dz \, p(z, y) \log \frac{p(z|y)}{p(z)}
\end{aligned}
\end{equation}
Then, we can derive 
\begin{equation}
\begin{aligned}
I(Z, T) &= \int dtdz p(z,t) \log \frac{p(t|z)}{p(t)} \\
&=\int dtdz p(z,t) \log \frac{q(t|z)}{p(t)} \frac{p(t|z)}{q(t|z)} \\
&= \int dtdz p(z,t) \log \frac{q(t|z)}{p(t)} + KL(p(t|z) || q(t|z)) \\
&\geq \int dtdz p(z,t) \log \frac{q(t|z)}{p(t)} \\
&= \int dtdz p(z,t) \log q(t|z) - \int dtdz p(z,t) \log p(t) \\
&= \int dtdz p(z,t) \log q(t|z) + H(t)
\end{aligned}
\end{equation}
Notice that $H(t)$ is independent in our optimization procedure and thus we can derive:
\begin{equation}
\begin{aligned}
I(Z,T) &\geq \int dtdz p(z,t) \log q(t|z) \\
&= \int dydtdz p(y)p(z|y)p(t|y) \log q(t|z)
\end{aligned}
\label{eq:IZT}
\end{equation}
For $I(Z, Y)$, we have
\begin{equation}
\begin{aligned}
I(Z, Y) &= \int dydz p(z,y) \log \frac{p(y|z)}{p(y)} \\
&=\int dtdz p(z,y) \log \frac{r(z)}{p(z)} \frac{p(z|y)}{r(z)} \\
&= -KL(p(z) || r(z)) + \int dydz p(z,y) \log \frac{p(z|y)}{r(z)} \\
&\leq \int dydz p(z,y) \log \frac{p(z|y)}{r(z)} \\
&= \int dydz p(y)p(z|y) \log \frac{p(z|y)}{r(z)}
\end{aligned}
\label{eq:IZY}
\end{equation}
By the combination of \cref{eq:IZT} and \cref{eq:IZY}, we have
\begin{equation}
\begin{aligned}
I(Z, T) - \beta I(Z, Y) \geq
\int dydtdz p(y)p(z|y)p(t|y) \log q(t|z) \\
-\beta \int dydz p(y)p(z|y) \log \frac{p(z|y)}{r(z)}\\
\approx \int dz p(z|y) \log q(t|z) - \beta p(z|y) \log \frac{p(z|y)}{r(z)}
\end{aligned}
\end{equation}
Finally, the objective function for VIB can be denoted as: 
\begin{equation}
\begin{aligned}
\mathcal{L}_{IB} = \mathbb{E}_{z \sim p(z|y)}[-\log q(t|z)] \\
+ \beta KL(p(z|y) || r(z))
\label{eq:loss_ib}
\end{aligned}
\end{equation}
where $q(t|z)$ is the variational approximation to $p(t|z)$, $r(z)$ is the variational approximation of $p(z)$ and $p(z|y)$is the posterior distribution over $z$. 


\subsection{Conditional Latent Differentiation VAE}
\subsubsection{Latent Differentiation VAE}
In general, we assume that the $N$ functional genomic features $x_1, x_2,...,x_N$ are conditionally independent given a latent variable $z_X$. Consequently, the objective of training this VAE is to maximize the likelihood of the data $p(X)=p({x_1,x_2,_,...,x_N})$, which can be optimized using an evidence lower bound (ELBO), and the loss function can be defined as:
\begin{equation}
\begin{aligned}
\mathcal{L}_{ELBO} = -\mathbb{E}_{q_{\phi}(z_X|x)}[\sum_{i=1}^{N} \log p_{\theta}(x_i|z_X)] \\
+ \beta KL[q_{\phi}(z_X|X), p(z_X)]
\end{aligned}
\end{equation}
However, it is difficult to generate different functional genomic features $x_i$ directly from the genomic posterior $p(z|x_1, x_2,...,x_N)$ as the genomic posterior will affect the diversity of the generated genomic features~\cite{wu2018multimodal}. To address this, we introduce the function-specific posteriors $p(z_{i}|X)$ by applying a latent differentiation process that transforms the genomic posterior into function-specific posteriors. Therefore, we can establish a shared latent space on all genomic features as well as refining the function-specific posteriors for each genomic category.

Specifically, we assign a unique latent variable $z_{i}$ to each $x_i$ and assume that $x_i$, $z_{i}$ and $z_X$ satisfy the Markov chain $x_i \leftrightarrow z_{i} \leftrightarrow z_{X}$. Due to the one-to-one correspondence, we have $p(x_i|z_X) = p(x_i,z_{i}|z_X)$. Then the generative model is with the following form:
\begin{equation}
\begin{aligned}
p(x_1,z_{1},x_2,z_{2},..., x_N, z_{N},z_X)=p(x_1, x_2, ..., x_{N}, z) \\
= p(z_X) p(x_1|z_X) p(x_2|z_X) ... p(x_N|z_X)
\end{aligned}
\end{equation}
We assume that $p(z_{i}|X)=\mathbb{E}_{z_X \sim p(z_X|X)}p(z_{i}|z_X)$, indicating that $p(z_{i}|X)$ can be obtained by transforming from the genomic posterior. Since $p(z_X|X)$ is variationally approximated by $q(z_X|X)$, our function-specific posterior can be derived directly from the variational genomic posterior, avoiding the need for an independent variational approximation of $q(z_X|X)$. Therefore, we have $p(z_{i}|X) \approx \mathbb{E}_{z_X \sim q(z_X|X)} p(z_{i} | z_X)$. We model $p(z_{i}|z_X)$ as a process that maps $p(z_X)$ to $p(z_{i})$, which can be presented as $p(z_{i}|z_X)=\mathcal{N}(z_{i}|\psi_{i}^{\mu}(z_X), \psi_{i}^{\Sigma}(z_X))$,where $\psi_i$ serves as an MLP mapper. In practice, each function-specific posterior should also approximate prior distribution $r(z_{i})$, which can be achieved by applying the Kullback-Leibler divergence.

In the process for variational inference, we sample the genomic latent variable $z_X$ from the prior $p(z_X) \sim \mathcal{N}(0,I)$. Then, the joint distribution $p(x_i, z_{i}, z_X)$ can be factored as follows:
\begin{equation}
\begin{aligned}
p(x_i,z_{i},z_X) &= p(x_i|z_{i},z_X) p(z_{i}|z_X) p(z_X) \\
&= p(x_i|z_{i}) p(z_{i}|z_X) p(z_X)
\end{aligned}
\end{equation}
Here, $p(x_i|z_{i},z_X)=p(x_i|z_{i})$ follows the Markov chain. Then the $\log p_{\theta}(x_i|z_X)$ in \cref{eq:elbo_ldvae1} can be reformulated as:
\begin{equation}
\log p(x_i|z_X) = \log p(x_i, z_{i} | z_X) 
= \log p(x_i | z_{i}) p(z_{i}|z_X)
\end{equation}
By combining the above analysis, the loss function for LD-VAE is denoted as:
\begin{equation}
\begin{aligned}
\mathcal{L}_{ELBO} = -\mathbb{E}_{q_{\phi}(z_X|X)}[\sum_{i=1}^{N} \log p_{\theta}(x_i|z_{i}) p(z_{i}|z_X)] \\
+ \beta (\sum^{N}_{i=1} KL[q_{\phi}(z_{i}|X) || p(z_{i})] + KL[q_{\phi}(z_X|X) || p(z_X)])
\end{aligned}
\end{equation}

\paragraph{The architecture of LD-VAE.}
We show the detailed architecture of LD-VAE in \cref{fig:architecture}(b). Specifically, we employ a transformer similar to the VIB-Trans as the encoder. The transformer encoder takes the bag of genomic feature $X=\{X_1, X_2, ..., X_{N}\}$ as input with two additional learnable tokens, $\mu_{X}^{token}$ and $\Sigma_{X}^{token}$, and outputs the parameters $\mu_X$ and $\Sigma_X$ of the genomic posterior distribution in LD-VAE. 
For the reconstruction of genomic features, we set specific decoders for the genomic features with diverse biological functions. The specific decoder first uses the mapper $\psi_i$ to generate the function-specific posterior from the genomic posterior, and then obtain the specific latent variable $z_{i}$  with the re-parametrization trick to generate the genomic features $x_i$ by the reconstruction net $\theta_i$.

\subsubsection{Joint Pathology-Genomics Distribution Learning}
One critical problem for optimizing \cref{eq:elbo_ldvae3} is to estimate the joint posterior $q(z|X,Y)$. Following the study in \cite{wu2018multimodal}, we assume that $X$,$Y$ are conditionally independent given the genomic latent variable z, \emph{i.e.,} $p(X,Y|z)=p(X|z)p(Y|z)$. Hence, the joint posterior can be approximated by the product of the genomics and pathology posteriors with the form:
\begin{equation}
\begin{aligned}
p(z|X,Y) &= \frac{p(X,Y|z) p(z)}{p(X,Y)} \\ &=\frac{p(z)}{p(X,Y)} p(X|z) p(Y|z) \\
&= \frac{p(z)}{p(X,Y)} \frac{p(z|X) p(X)}{p(z)} \frac{p(z|Y) p(Y)}{p(z)} \\
&= \frac{p(z|X) p(z|Y)}{p(z)} \frac{p(X) p(Y)}{p(z)} \\
&\propto \frac{p(z|X) p(z|Y)}{p(z)}
\end{aligned}
\end{equation}
By the approximation of $p(z|X) \equiv \tilde{q}(z|X) p(z)$, $p(z|Y) \equiv \tilde{q}(z|Y) p(z)$, where $\tilde{q}(\cdot)$ is the underlying inference network, we can derive:
\begin{equation}
\begin{aligned}
p(z|X,Y) \propto \frac{p(z|X) p(z|Y)}{p(z)} \\
\approx \tilde{q}(z|X) \tilde{q}(z|Y) p(z) \equiv q(z|X, Y)
\end{aligned}
\end{equation}
Finally, we can use the product-of-experts (PoE) that factorizes the joint posterior $q(z|X,Y)$ into  marginal posteriors $\tilde{q}(z|X)$ and $\tilde{q}(z|Y)$.

\begin{figure*}
    \centering
    \includegraphics[width=1.0\linewidth]{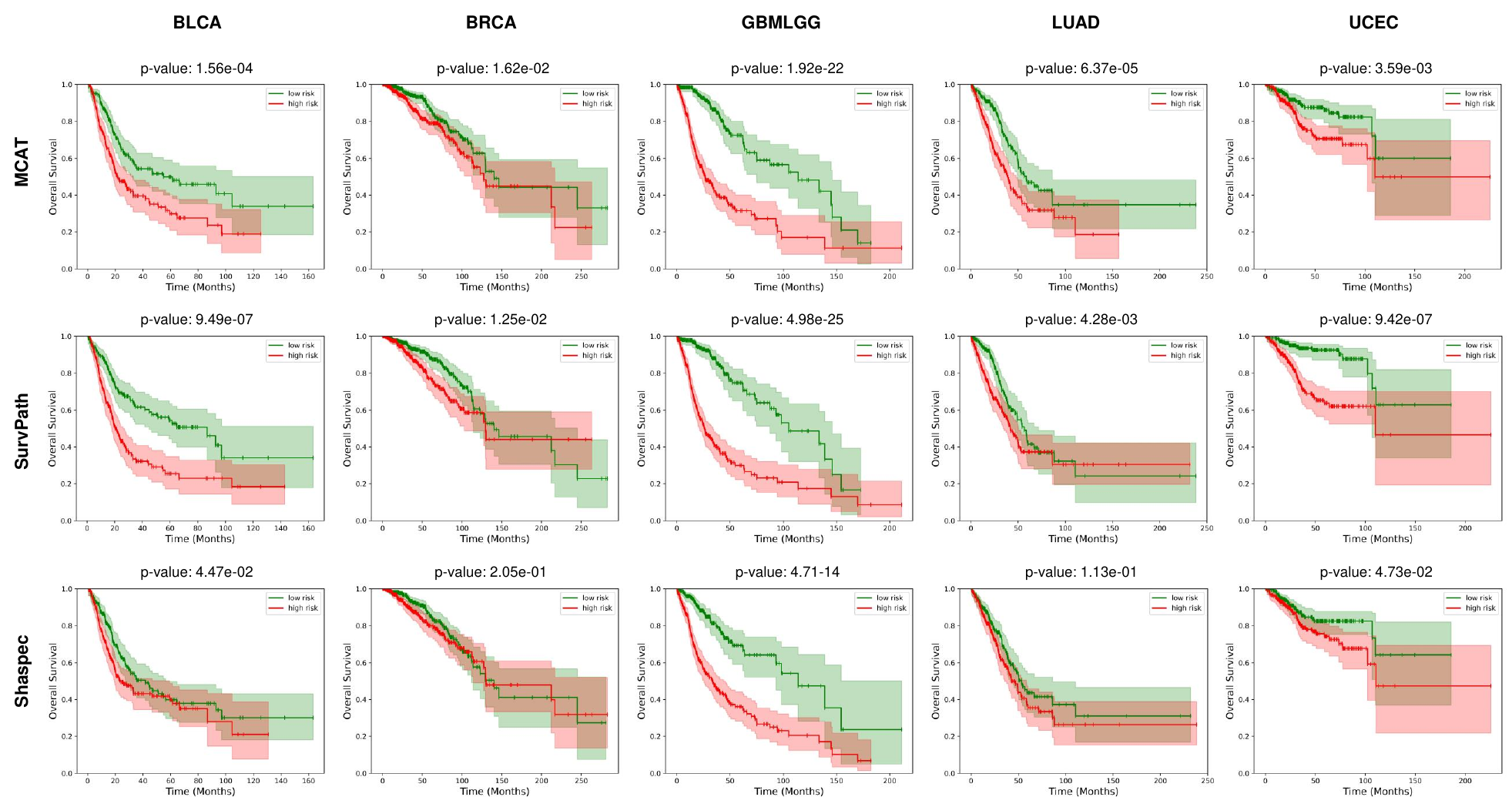}
    \caption{Kaplan-Meier Analysis of predicted high-risk (red) and low-risk (green) groups on five cancer datasets under both complete modality (top) and missing modality (bottom) scenarios. Shaded areas refer to the confidence intervals.}
    \label{fig:exp_curves_others}
\end{figure*}

\begin{table*}[htb]
\setlength{\arrayrulewidth}{0.1mm}  
\tiny 
\centering
\caption{Comparisons of C-Index (mean ± std) with different methods for the representation of WSIs over five datasets. For each method, results are reported under both complete modality and missing modality scenarios.}
\label{suptable: wsirepresentation}
\resizebox{\linewidth}{!}{
\begin{tabular}{cccccccc}
\hline
\hline
\textbf{Model} & \textbf{Missing}  & \textbf{BLCA} & \textbf{BRCA} & \textbf{GBMLGG} & \textbf{LUAD} & \textbf{UCEC} & \textbf{Overall} \\
\hline
\hline
\multirow{2}{*}{{ABMIL}} & & 0.651 \texttt{\textpm}	0.035 & 0.685 \texttt{\textpm}	0.020 &	0.833 \texttt{\textpm}	0.058 &	0.646 \texttt{\textpm}	0.048 &	0.691 \texttt{\textpm}	0.053 &	0.701 \\
& \checkmark & 0.616 \texttt{\textpm}	0.024 &	0.619 \texttt{\textpm}	0.019 &	0.820 \texttt{\textpm}	0.040 &	0.623 \texttt{\textpm}	0.035 &	0.647 \texttt{\textpm}	0.037 &	0.665 \\
\hline
\multirow{2}{*}{{CLAM-SM}} & & 0.651 \texttt{\textpm}	0.035 & 0.685 \texttt{\textpm}	0.020 &	0.833 \texttt{\textpm}	0.058 &	0.646 \texttt{\textpm}	0.048 &	0.691 \texttt{\textpm}	0.053 &	0.701 \\
& \checkmark & 0.616 \texttt{\textpm}	0.024 &	0.619 \texttt{\textpm}	0.019 &	0.820 \texttt{\textpm}	0.040 &	0.623 \texttt{\textpm}	0.035 &	0.647 \texttt{\textpm}	0.037 &	0.665 \\
\hline
\multirow{2}{*}{{CLAM-MB}} & & 0.651 \texttt{\textpm}	0.035 & 0.685 \texttt{\textpm}	0.020 &	0.833 \texttt{\textpm}	0.058 &	0.646 \texttt{\textpm}	0.048 &	0.691 \texttt{\textpm}	0.053 &	0.701 \\
& \checkmark & 0.616 \texttt{\textpm}	0.024 &	0.619 \texttt{\textpm}	0.019 &	0.820 \texttt{\textpm}	0.040 &	0.623 \texttt{\textpm}	0.035 &	0.647 \texttt{\textpm}	0.037 &	0.665 \\
\hline
\multirow{2}{*}{{TransMIL}} & & 0.651 \texttt{\textpm}	0.035 & 0.685 \texttt{\textpm}	0.020 &	0.833 \texttt{\textpm}	0.058 &	0.646 \texttt{\textpm}	0.048 &	0.691 \texttt{\textpm}	0.053 &	0.701 \\
& \checkmark & 0.616 \texttt{\textpm}	0.024 &	0.619 \texttt{\textpm}	0.019 &	0.820 \texttt{\textpm}	0.040 &	0.623 \texttt{\textpm}	0.035 &	0.647 \texttt{\textpm}	0.037 &	0.665 \\
\hline
\multirow{2}{*}{{VIB-Trans}}  & & 0.686  \texttt{\textpm} 0.035  & 0.680  \texttt{\textpm} 0.030  & 0.849  \texttt{\textpm} 0.017  & 0.676  \texttt{\textpm}0.015  & 0.703  \texttt{\textpm} 0.069  & 0.719 \\
 & \checkmark  & 0.649  \texttt{\textpm} 0.040  & 0.641  \texttt{\textpm} 0.012  & 0.821  \texttt{\textpm} 0.021  & 0.628  \texttt{\textpm} 0.008  & 0.681  \texttt{\textpm} 0.044  & 0.684 \\
\hline
\hline
\end{tabular}
}
\end{table*}

\begin{table*}[htb]
\setlength{\arrayrulewidth}{0.1mm}  
\tiny 
\centering
\caption{Ablation study assessing C-index (mean ± std) over five datasets. For each variant, results are reported under both complete modality and missing modality scenarios.}
\label{suptable: tokens}
\resizebox{\linewidth}{!}{
\begin{tabular}{cccccccc}
\hline
\hline
\textbf{Variants} & \textbf{Missing}  & \textbf{BLCA} & \textbf{BRCA} & \textbf{GBMLGG} & \textbf{LUAD} & \textbf{UCEC} & \textbf{Overall} \\
\hline
\hline
\multirow{2}{*}{w/o $\mu^{token}$ and $\Sigma^{token}$}  & & 0.675 \texttt{\textpm} 0.063 & 0.674 \texttt{\textpm} 0.033 & 0.839 \texttt{\textpm} 0.035 & 0.668 \texttt{\textpm} 0.039 & 0.682 \texttt{\textpm} 0.081 & 0.708 \\
& \checkmark  & 0.641 \texttt{\textpm} 0.040 &	0.644 \texttt{\textpm}	0.012 &	0.811 \texttt{\textpm}	0.027 &	0.629 \texttt{\textpm}	0.022 &	0.643 \texttt{\textpm}	0.036 &	0.674 \\
\hline
\multirow{2}{*}{Ours}  & & {0.686  \texttt{\textpm} 0.035}  & {0.680  \texttt{\textpm} 0.030}  & {0.849  \texttt{\textpm} 0.017}  & {0.676  \texttt{\textpm}0.015}  & {0.703  \texttt{\textpm} 0.069}  & {0.719} \\
 & \checkmark  & 0.649  \texttt{\textpm} 0.040  & 0.641  \texttt{\textpm} 0.012  & 0.821  \texttt{\textpm} 0.021  & 0.628  \texttt{\textpm} 0.008  & 0.681  \texttt{\textpm} 0.044  & 0.684 \\
\hline
\hline
\end{tabular}
}
\end{table*}

\begin{figure*}
    \centering
    \includegraphics[width=1.0\linewidth]{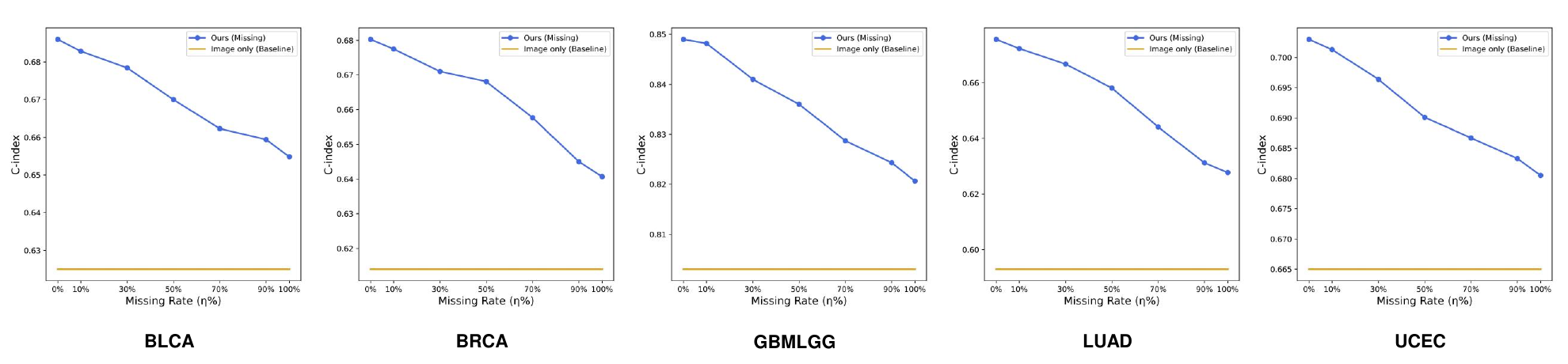}
    \caption{The performance of our method under different settings of missing rate $\eta$.}
    \label{fig:exp_missingrate}
\end{figure*}

\begin{figure}
    \vspace{-3mm}
    \centering
    \includegraphics[width=1.0\linewidth]{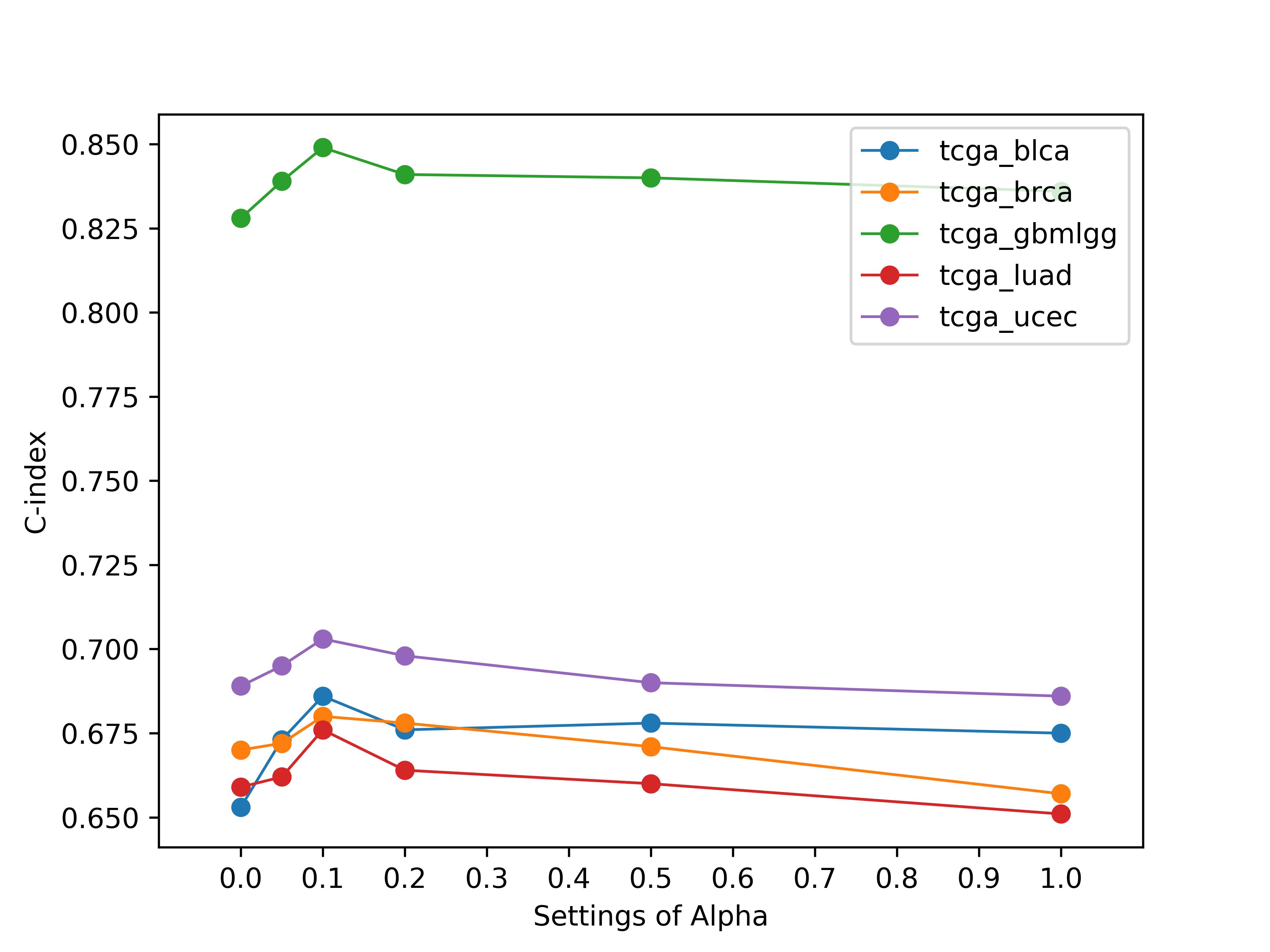}
    \caption{The effect of hyperparameter $\alpha$ over five cancer datasets.}
    \label{fig:hyperparameter}
    \vspace{-2mm}
\end{figure}

\section{Additional Experiments}
\subsection{Evaluation and Implementation}
\paragraph{Evaluation.} 
We employ 5-fold cross-validation for each dataset. To evaluate the model's performance, we calculate the concordance index (C-index) \cite{harrell1996multivariable} and its standard deviation (std), which measure the model’s ability to correctly rank pairs of individuals based on their predicted survival times. Additionally, we visualize the Kaplan-Meier (KM) \cite{kaplan1958nonparametric} survival curves to illustrate the survival probability of different risk groups predicted by our model. To statistically validate the separation between risk groups, we perform the Log-rank test \cite{mantel1966evaluation}, which determines whether the survival differences between groups are statistically significant.
\paragraph{Implementation.}
For each WSI, we crop it into non-overlapping 224$\times$224 patches at 10$\times$magnification level. Then, a pre-trained Swin Transformer encoder (\emph{i.e.,} CTransPath) serves as the pathological encoder $\varphi_{p}$, which is pre-trained using contrastive learning on over 15 million pan-cancer histopathology patches \cite{wang2021transpath, wang2022transformer}.
For genomic data, we organizes genes into the aforementioned N=6 categories based on similar biological functional impact, which are obtained from \cite{liberzon2015molecular}. We use the SNN \cite{klambauer2017self} as the genomic encoder $\varphi_{g}$. 
Our model is implemented in Python 3.9 with Pytorch library and trained with four NVIDIA 3090 GPUs. During training, we set the hyperparameter $\alpha$ to 0.1. The hyperparameter $\beta$ for Kullback-Leibler divergence is annealed using a cosine schedule, gradually increasing to 1, thereby forming a valid lower bound on the evidence. We adopt Adam optimizer with the initial learning rate of 2$\times$10$^{-4}$ and weight decay of 1$\times$10$^{-5}$. Following the setting of \cite{chen2021multimodal}, we use the batch size of 1 for WSIs with 32 gradient accumulation steps, and all experiments are trained for 30 epoches. We train our methods on the complete data and test the performance on both the complete modality and missing modality (\emph{i.e.,} genomic data).

\subsection{Additional Results}
\paragraph{Additional Results for Patient Stratification.} We present additional results for patient stratification using competing methods, including those designed for integrating multi-modal data (\emph{e.g.,} MCAT, SurvPath) and handling missing genomic data (\emph{e.g.,} Shaspace), as shown in \cref{fig:exp_curves_others}. Compared to these methods, our approach (shown in \cref{fig:exp_curves_ours}) achieves clearer separation between low-risk and high-risk patients across all datasets. In the Logrank test, our method can still consistently yield a lower P-value. These results highlight the robustness and effectiveness of our approach in accurately distinguishing patient risk groups.

\paragraph{Representation of WSIs.}
We conduct additional experiments with different methods to represent WSIs in place of VIB-Trans, with the results summarized in \cref{suptable: wsirepresentation}. We can observe that VIB-Trans consistently outperforms competing methods across all datasets, demonstrating its effectiveness of representing WSIs.

\paragraph{Settings of missing rate for genomic data.}
We conduct further experiments to analyze the robustness of our proposed method against different missing rates for genomic data, and the results are shwon in \cref{fig:exp_missingrate}. We observe that the increase of missing rate $\eta$ comes with inferior prediction results. However, even at high missing rates of 90\% or even 100\%, our method still outperforms the best uni-modal approach basing on pathology images, indicating the effectiveness of our approach.

\paragraph{Settings of hyperparameters for Loss Function.}
We conduct ablation studies to evaluate the impact of the hyperparameter $\alpha$ in \cref{eq:ldcvae}. As shown in \cref{fig:hyperparameter}, model performance peaks when $\alpha$ is 0.1, hence we set 0.1 as the optimal value of $\alpha$.

\paragraph{Impact of $\mu^{token}, \sigma^{token}$.}
We introduce the $\mu^{token}$ and $\Sigma^{token}$in Trans-VIB to better learn the latent posterior of VIB by modeling the interaction between pairwise patches. For ablation, we use two linear layers following the transformer encoder to obtain the $\mu$ and $\Sigma $. As shown in \cref{suptable: tokens}, the inclusion of  $\mu^{token}$ and $\Sigma^{token}$ achieves superior performance, indicating the effectiveness of this design.

\subsection{More Comparisons with State-of-the-Arts}
We also conduct additional experiments with pathological features extracted by a ResNet50 encoder pre-trained on ImageNet, and report the results with comparisons to SOTA methods reported in \cref{suptable:resnet50_1,suptable:resnet50_2}. The results in \cref{suptable:resnet50_1} demonstrate that our method consistently achieves the best overall performance across both unimodal and multimodal approaches. In the missing modality setting, results in \cref{suptable:resnet50_2} show that our method can effectively handle missing genomic data and outperform the comparison methods.

\subsection{More Visualizations}
\begin{figure*}
    \centering
    \includegraphics[width=1.0\linewidth]{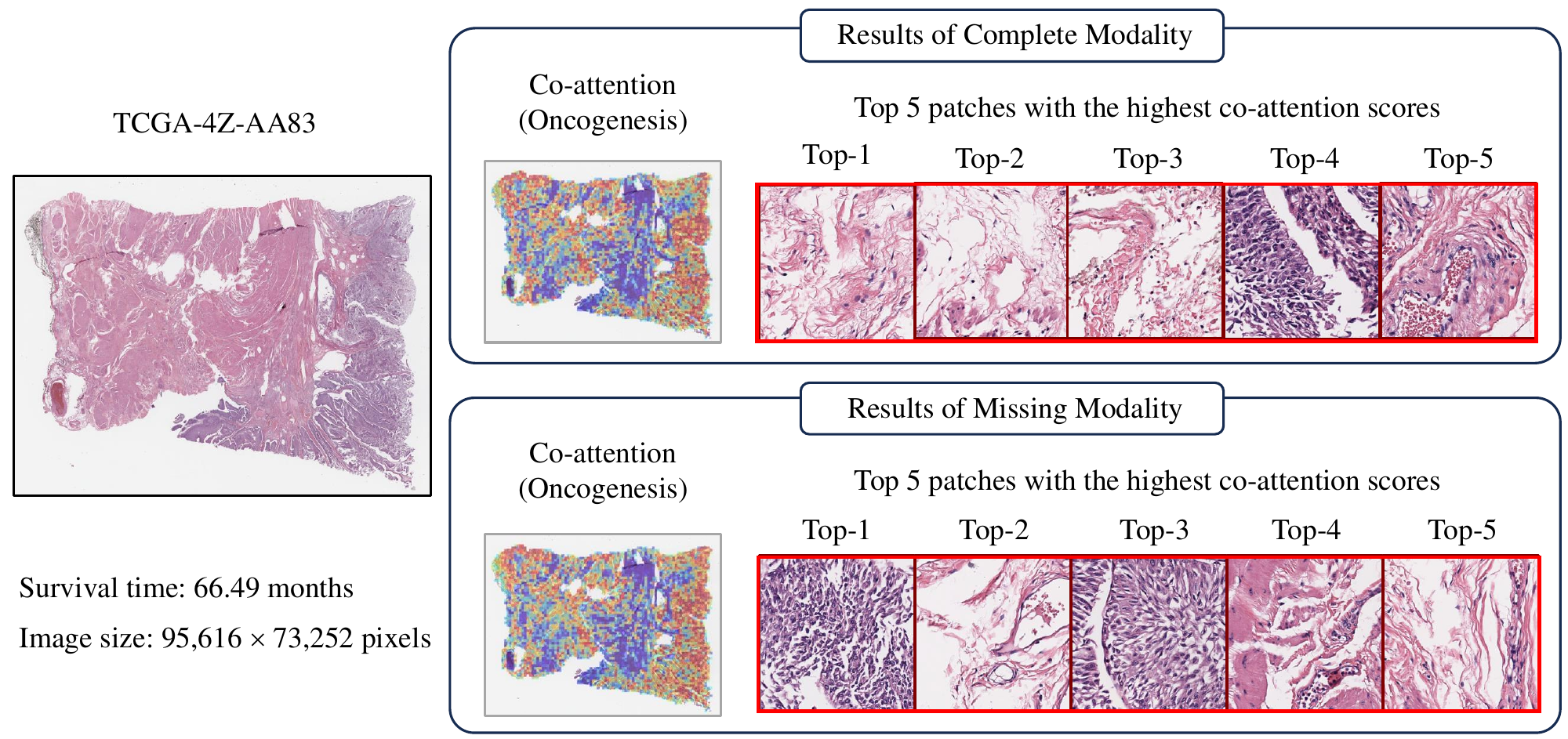}
    \caption{Comparison of the co-attention weights calculated from the genuine (top) and generated (bottom) genomic features.}
    \label{fig:supfig_visual1}
\end{figure*}

\begin{figure*}
    \centering
    \includegraphics[width=1.0\linewidth]{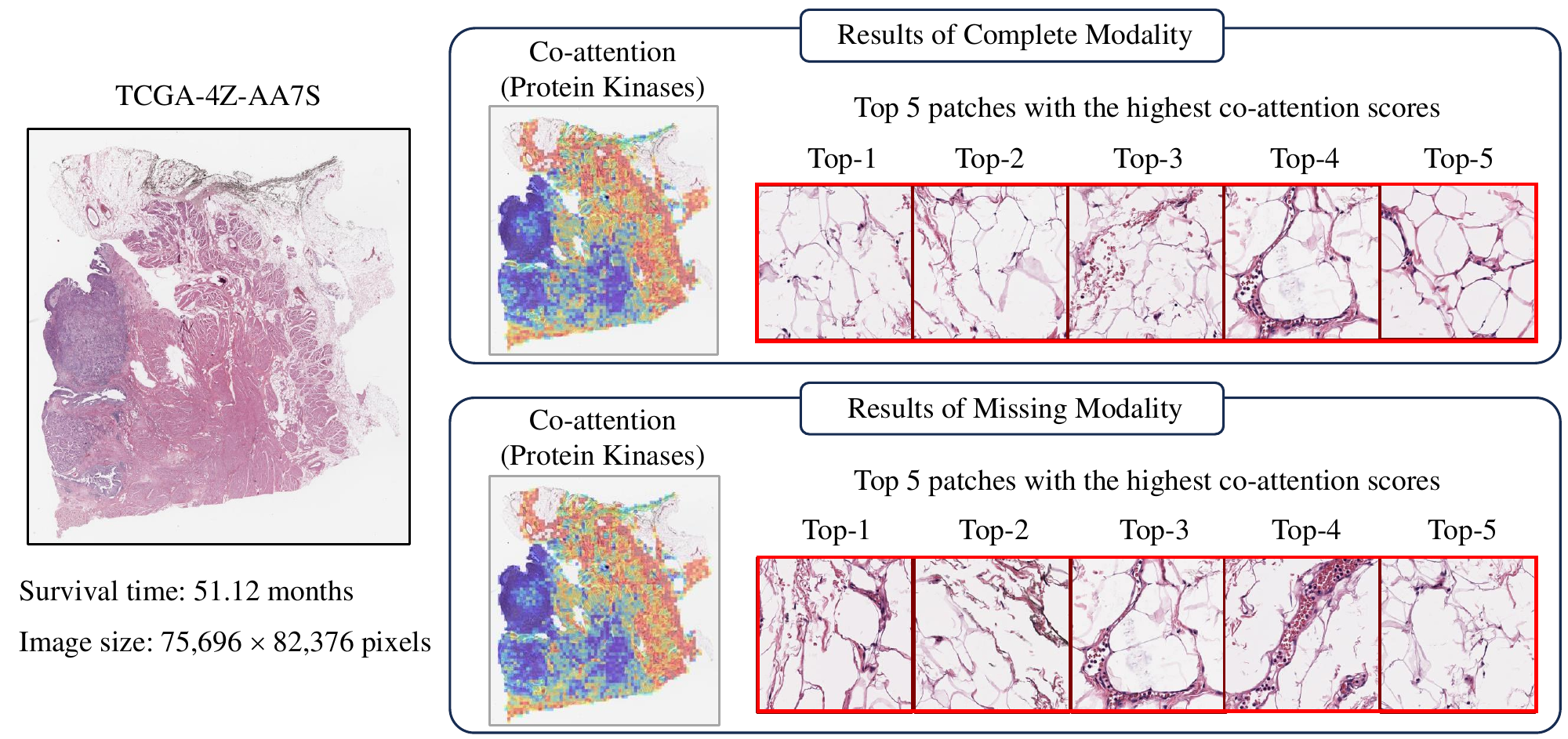}
    \caption{Comparison of the co-attention weights calculated from the genuine (top) and generated (bottom) genomic features.}
    \label{fig:supfig_visual2}
\end{figure*}

\begin{figure*}
    \centering
    \includegraphics[width=1.0\linewidth]{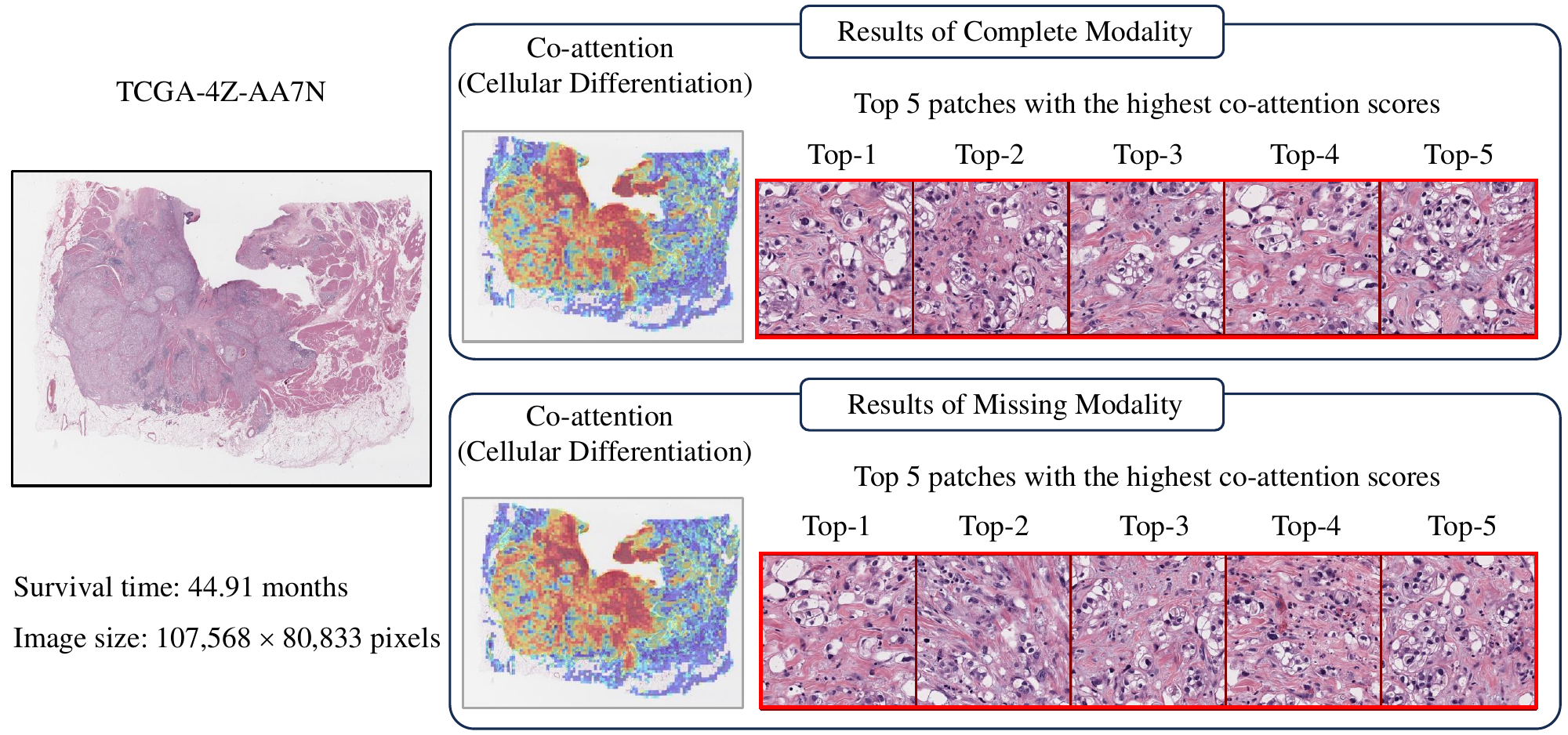}
    \caption{Comparison of the co-attention weights calculated from the genuine (top) and generated (bottom) genomic features.}
    \label{fig:supfig_visual3}
\end{figure*}

\begin{figure*}
    \centering
    \includegraphics[width=1.0\linewidth]{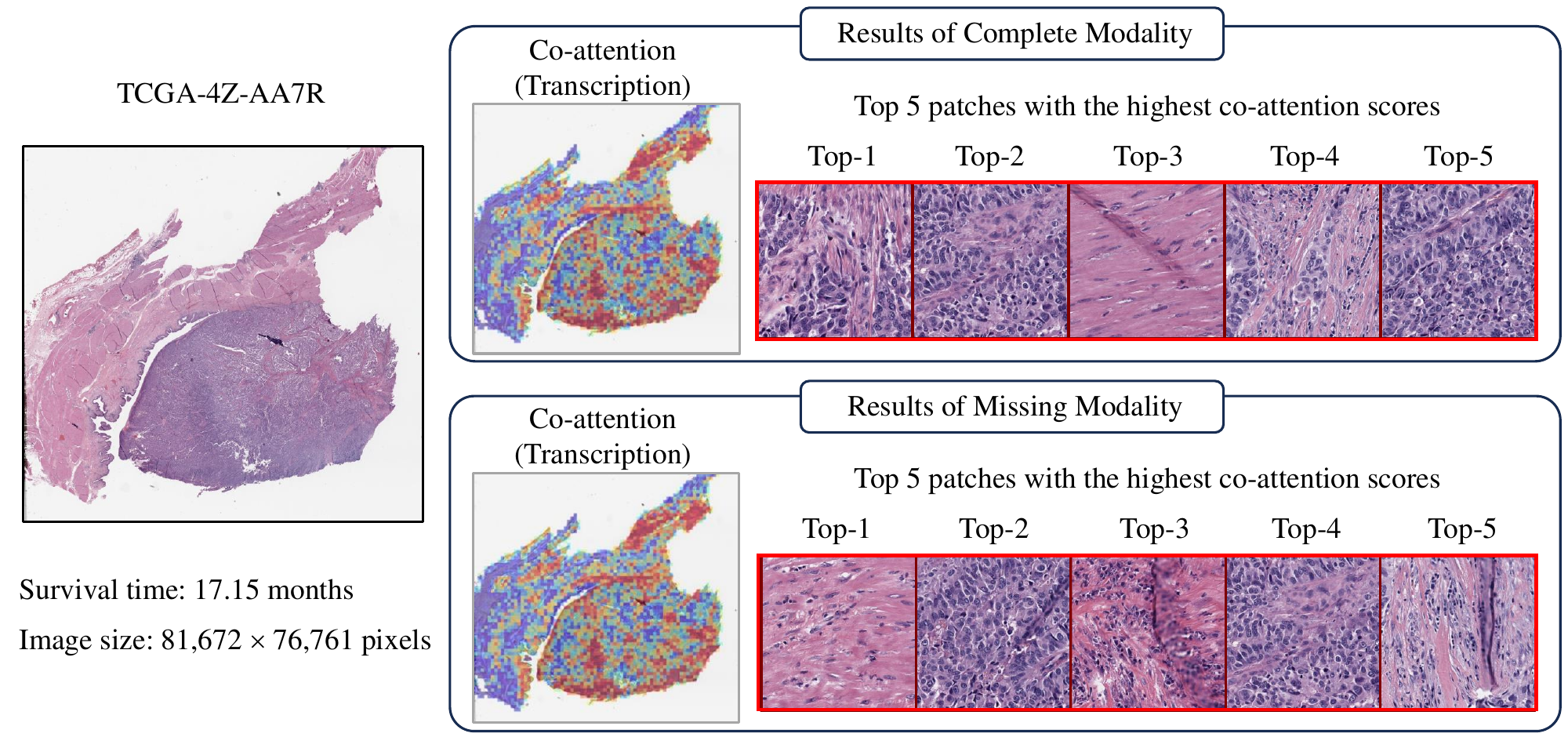}
    \caption{Comparison of the co-attention weights calculated from the genuine (top) and generated (bottom) genomic features.}
    \label{fig:supfig_visual4}
\end{figure*}

\begin{figure*}
    \centering
    \includegraphics[width=1.0\linewidth]{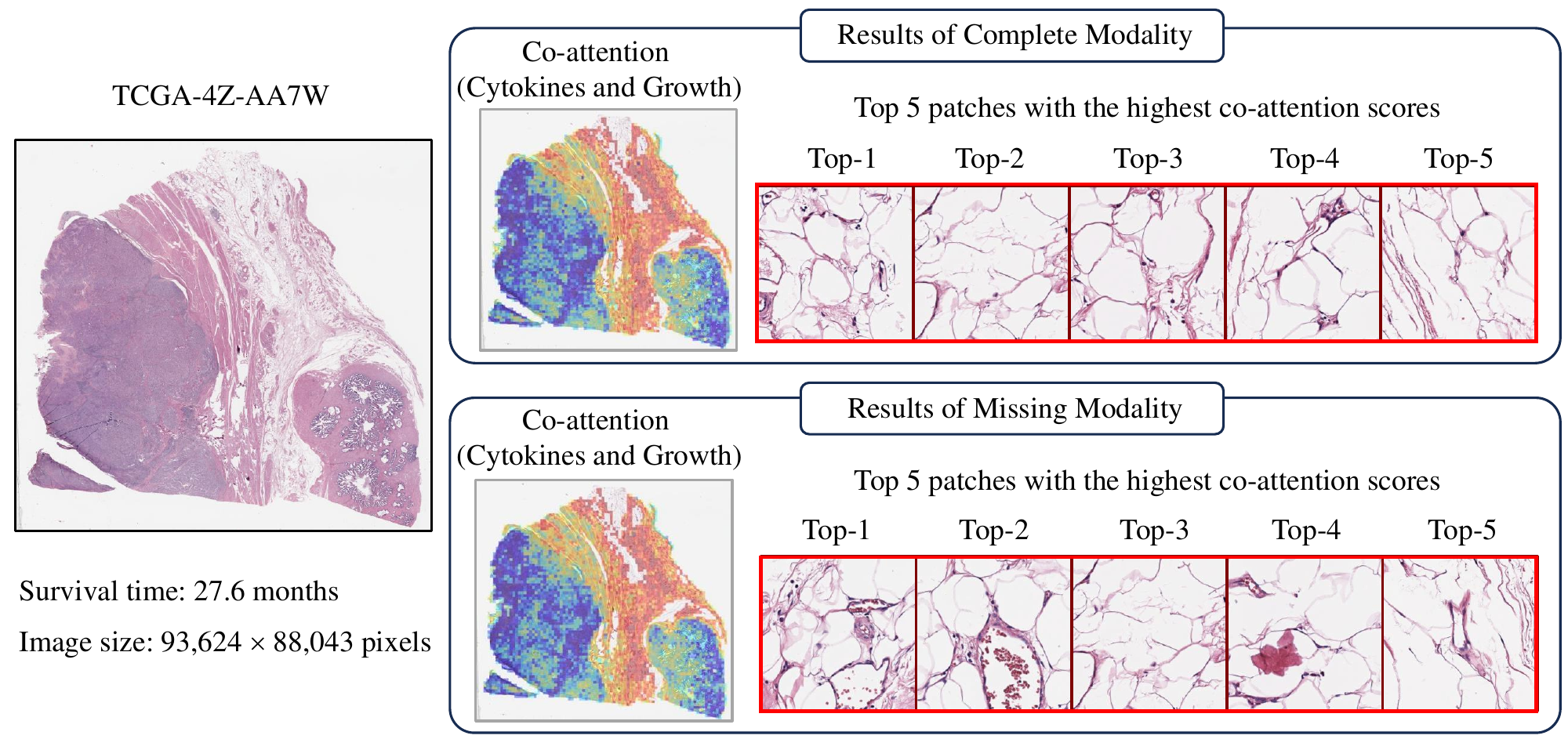}
    \caption{Comparison of the co-attention weights calculated from the genuine (top) and generated (bottom) genomic features.}
    \label{fig:supfig_visual5}
\end{figure*}

We provide more visualizations that compare the co-attention weights calculated from the ground truth and generated genomic feature in \cref{fig:supfig_visual1,fig:supfig_visual2,fig:supfig_visual3,fig:supfig_visual4,fig:supfig_visual5}.

\begin{table*}[htbp]
\setlength{\arrayrulewidth}{0.1mm}  
\tiny
\centering
\caption{Comparisons of C-index (mean ± std) with SOTA methods over five cancer datasets by using ResNet50 encoder. g. and h. refer to genomic modality and histological modality, respectively. The best results and the second-best results are highlighted in \textbf{bold} and in \underline{underline}.}
\label{suptable:resnet50_1}
\resizebox{\linewidth}{!}{
\begin{tabular}{lccccccc}
\hline
\hline
\multirow{2}{*}{\textbf{Model}}  & \multirow{2}{*}{\textbf{Modality}} & \textbf{BLCA} & \textbf{BRCA} & \textbf{GBMLGG} & \textbf{LUAD} & \textbf{UCEC} & \multirow{2}{*}{\textbf{Overall}} \\
 &  & (N=373) & (N=957) & (N=571) & (N=452) & (N=480) &  \\
\hline
\hline
MLP & g. & 0.613   \texttt{\textpm} 0.019  & 0.587 \texttt{\textpm} 0.033  & 0.809 
\texttt{\textpm} 0.029  & 0.617  \texttt{\textpm} 0.026  & 0.657  \texttt{\textpm} 0.036  & 0.657 \\
SNN & g. & 0.619  \texttt{\textpm} 0.023  & 0.596  \texttt{\textpm} 0.027  & 0.805 \texttt{\textpm} 0.030   & 0.625  \texttt{\textpm} 0.019  & 0.651  \texttt{\textpm} 0.018  & 0.659  \\
SNNTrans & g. & 0.627 \texttt{\textpm} 0.019 & 0.618  \texttt{\textpm} 0.018 & 0.816  \texttt{\textpm} 0.037  & 0.631  \texttt{\textpm} 0.023  & 0.641 \texttt{\textpm} 0.026 & 0.667 \\
\hline
ABMIL & h. & 0.594 \texttt{\textpm} 0.033  & 0.601  \texttt{\textpm} 0.033  & 0.779  \texttt{\textpm} 0.035  & 0.579  \texttt{\textpm} 0.070  & 0.637  \texttt{\textpm} 0.024 & 0.638 \\
CLAM-SB & h. & 0.594 \texttt{\textpm} 0.047& 	0.595 \texttt{\textpm}	0.028& 	0.787 \texttt{\textpm}	0.036& 	0.580 \texttt{\textpm}	0.053& 	0.648 \texttt{\textpm}	0.032& 	0.641  \\
CLAM-MB & h. & 0.598 \texttt{\textpm} 0.030 &	0.600 \texttt{\textpm}	0.017 &	0.790 \texttt{\textpm}	0.031 & 0.582 \texttt{\textpm}	0.077 &	0.657 \texttt{\textpm}	0.038 &	0.645  \\
TransMIL & h. & 0.605 \texttt{\textpm} 0.054 &	0.604 \texttt{\textpm}	0.054 &	0.793 \texttt{\textpm}	0.028 &	0.590 \texttt{\textpm}	0.057 &	0.649 \texttt{\textpm}	0.053 &	0.648  \\
\hline
Porpoise & g.+h. & 0.646 \texttt{\textpm} 0.038 & 0.652 \texttt{\textpm}	0.022 &	0.819 \texttt{\textpm}	0.033 &	0.649 \texttt{\textpm}	0.030 &	0.665 \texttt{\textpm}	0.043 &	0.685 \\
MCAT & g.+h. & 0.645 \texttt{\textpm} 0.031  &	0.648 \texttt{\textpm}	0.011 &	0.826 \texttt{\textpm}	0.033 &	0.651 \texttt{\textpm}	0.043 &	0.659 \texttt{\textpm}	0.062 &	0.690 \\
MOTCat & g.+h. & 0.649 \texttt{\textpm} 0.016 & 0.646 \texttt{\textpm} 0.055 & 0.829 \texttt{\textpm} 0.039 &	0.654 \texttt{\textpm} 0.031 &	0.651 \texttt{\textpm} 0.053 &	0.687 \\
CMTA & g.+h. & 0.653 \texttt{\textpm} 0.035 &	0.656 \texttt{\textpm}	0.045 &	\textbf{0.837 \texttt{\textpm} 0.028} & 0.657 \texttt{\textpm}	0.029 &	0.660 \texttt{\textpm}	0.035 &	0.693 \\
SurvPath & g.+h. & \underline{0.651\texttt{\textpm}0.028} & \underline{0.667  \texttt{\textpm} 0.053} & 0.833 \texttt{\textpm}0.043 & \underline{0.660 \texttt{\textpm} 0.015} & \underline{0.674 \texttt{\textpm} 0.051} & \underline{0.697}  \\
PIBD & g.+h. & 0.611 \texttt{\textpm}	0.012 &	0.606 \texttt{\textpm}	0.020 &	0.783 \texttt{\textpm}	0.056 &	0.621 \texttt{\textpm} 0.013 &	0.632 \texttt{\textpm}	0.038 & 0.651  \\
\rowcolor{gray!20}
Ours & g.+h. & \textbf{0.678 \texttt{\textpm}	0.048} & \textbf{0.668\texttt{\textpm}0.042} & \underline{0.835 \texttt{\textpm} 0.037} & \textbf{0.664 \texttt{\textpm} 0.035} & \textbf{0.680 \texttt{\textpm} 0.049} & \textbf{0.703} \\
\hline
\hline
\end{tabular}
}
\end{table*}

\begin{table*}[htb]
\tiny
\centering
\caption{Comparisons of C-index (mean ± std) with methods addressing missing modality over five cancer datasets by using ResNet50 encoder. The best results and the second-best results are highlighted in \textbf{bold} and in \underline{underline}.}
\label{suptable:resnet50_2}
\resizebox{\linewidth}{!}{
\begin{tabular}{ccccccc}
\hline
\hline
\multirow{2}{*}{\textbf{Model}}   & \textbf{BLCA} & \textbf{BRCA} & \textbf{GBMLGG} & \textbf{LUAD} & \textbf{UCEC} & \multirow{2}{*}{\textbf{Overall}} \\
 & (N=373) & (N=957) & (N=571) & (N=452) & (N=480) &  \\
\hline
\hline
VAE  & 0.588 \texttt{\textpm}	0.016 &	0.618 \texttt{\textpm}	0.028 &	0.788 \texttt{\textpm}	0.019 &	0.588 \texttt{\textpm}	0.044 &	0.636 \texttt{\textpm} 0.034 &	0.643\\
GAN  & 0.585 \texttt{\textpm}	0.011 & 0.611 \texttt{\textpm}	0.026 &	0.779 \texttt{\textpm}	0.029 & 0.599 \texttt{\textpm}	0.051 &	0.637 \texttt{\textpm}	0.044 &	0.642 \\
MVAE  & 0.588 \texttt{\textpm}	0.028 &	0.612 \texttt{\textpm}	0.029 &	0.774 \texttt{\textpm} 0.015 &	0.601 \texttt{\textpm}	0.026 &	0.636 \texttt{\textpm}	0.024 &	0.642 \\
SMIL  & 0.610 \texttt{\textpm}	0.020 &	0.615 \texttt{\textpm}	0.015 &	0.775 \texttt{\textpm}	0.021 &	0.599 \texttt{\textpm}	0.024 &	0.647 \texttt{\textpm}	0.024 &	0.649  \\
ShaSpec& \underline{0.615 \texttt{\textpm} 0.017} & 0.618 \texttt{\textpm}	0.028 &	0.791 \texttt{\textpm} 0.011 & \underline{0.611 \texttt{\textpm}	0.040} & \underline{0.656 \texttt{\textpm} 0.036} &	\textbf{0.658} \\
Transformer  & 0.613 \texttt{\textpm}	0.033 &	\underline{0.620 \texttt{\textpm}	0.022} & \underline{0.797 \texttt{\textpm} 0.022} & 0.602 \texttt{\textpm} 0.039 & 0.652 \texttt{\textpm} 0.013 & 0.657 \\
\rowcolor{gray!20}
Ours  & \textbf{0.637\texttt{\textpm}0.028} &	\textbf{0.634 \texttt{\textpm} 0.033} & \textbf{0.806 \texttt{\textpm} 0.026} & \textbf{0.634 \texttt{\textpm} 0.038} & \textbf{0.661 \texttt{\textpm} 0.029} & \textbf{0.672} \\

\hline
\hline
\end{tabular}
}
\end{table*}

\end{document}